\documentclass[a4paper,fleqn]{cas-dc}

\usepackage[numbers]{natbib}
\let\printorcid\relax % Remove ORCID footnote
\usepackage{hyperref}
\hypersetup{
    colorlinks=true,
    linkcolor=cyan,  
    urlcolor=cyan,
    citecolor=cyan,
}
\usepackage{mathtools}
\usepackage{accents}
\usepackage{amsmath}
\usepackage{amssymb}
\usepackage[labelformat=simple, font=small]{subcaption}

\usepackage{graphicx}
\usepackage{booktabs}
\usepackage{makecell}
\usepackage{multirow}
\usepackage{stfloats}
\usepackage[ruled,linesnumbered]{algorithm2e}
\usepackage[title]{appendix}

\usepackage{framed} % Framing content
\usepackage{multicol} % Multiple columns environment

\usepackage{nomencl} % Nomenclature package
\makenomenclature
\renewcommand*\nompreamble{\begin{multicols}{2}}
\renewcommand*\nompostamble{\end{multicols}}

\usepackage{etoolbox}

\newlength{\nomgroupstartsep}
\newcommand{\nomenclheader}[1]{\item[\hspace*{-\itemindent}\bfseries#1]}

\renewcommand\nomgroup[1]{%
 \itemsep\nomgroupstartsep%
  \IfStrEqCase{#1}{%
   {A}{\nomenclheader{Abbreviations}}% 
   {N}{\nomenclheader{Notations}}%
  }%
  \itemsep\nomitemsep% restore spacing
}

\def\tsc#1{\csdef{#1}{\textsc{\lowercase{#1}}\xspace}}
\tsc{WGM}
\tsc{QE}
\tsc{EP}
\tsc{PMS}
\tsc{BEC}
\tsc{DE}
\begin{document}
\let\WriteBookmarks\relax
\def\floatpagepagefraction{1}
\def\textpagefraction{.001}

\shorttitle{}

\shortauthors{C. Li et~al.}

\title [mode = title]{Optimal Power Flow in Highly Renewable Power System Based on Attention Neural Networks}                      
% \tnotemark[1,2]

% \tnotetext[1]{This document is the results of the research
%    project funded by the National Science Foundation.}

% \tnotetext[2]{The second title footnote which is a longer text matter
%    to fill through the whole text width and overflow into
%    another line in the footnotes area of the first page.}

\author[1,2]{Chen Li}


% \ead{chli@fias.uni-frankfurt.de}

% \ead[url]{www.cvr.cc, cvr@sayahna.org}

% \credit{Conceptualization of this study, Methodology, Software}

\affiliation[1]{organization={Frankfurt Institute for Advanced Studies, Goethe University Frankfurt},
    addressline={Ruth-Moufang Str. 1}, 
    city={Frankfurt am Main},
    citysep={}, % Uncomment if no comma needed between city and postcode
    postcode={60438}, 
    country={Germany}}

\affiliation[2]{organization={Xidian-FIAS International Joint Research Center, Xidian University},
    addressline={Taibai South Road 2}, 
    city={Xi'an, Shaanxi},
    citysep={}, % Uncomment if no comma needed between city and postcode
    postcode={710071}, 
    % state={Trivandrum},
    country={China}}

\author[1,3]{Alexander Kies}
\ead{kies@ece.au.dk}

\affiliation[3]{organization={Department of Electrical and Computer Engineering, Aarhus University},
    addressline={Nordre Ringgade 1}, 
    city={Aarhus C},
    citysep={}, % Uncomment if no comma needed between city and postcode
    postcode={8000}, 
    country={Denmark}}
    
\cormark[1]
\author[1]{Kai Zhou}
\ead{zhou@fias.uni-frankfurt.de}
\cormark[1]
% \ead{cvr3@sayahna.org}
% \ead[URL]{www.sayahna.org}

% \credit{Data curation, Writing - Original draft preparation}

\author[1]{Markus Schlott}
\author[1]{Omar El Sayed}
\author[1]{Mariia Bilousova}

\author[1,4]{Horst Stöcker}

\affiliation[4]{organization={GSI Helmholtzzentrum für Schwerionenforschung},
    addressline={Planckstraße 1}, 
    city={Darmstadt},
    citysep={}, % Uncomment if no comma needed between city and postcode
    postcode={64291}, 
    country={Germany}}
    
\cortext[cor1]{Corresponding authors}
% \cortext[cor2]{Principal corresponding author}

% \fntext[fn1]{This is the first author footnote. but is common to third
% \fntext[fn2]{Another author footnote, this is a very long footnote and
%   it should be a really long footnote. But this footnote is not yet
%   sufficiently long enough to make two lines of footnote text.}

% \nonumnote{This note has no numbers. In this work we demonstrate $a_b$
%   the formation Y\_1 of a new type of polariton on the interface
%   between a cuprous oxide slab and a polystyrene micro-sphere placed

\begin{abstract}
The Optimal Power Flow (OPF) problem is pivotal for power system operations, guiding generator output and power distribution to meet demand at minimized costs, while adhering to physical and engineering constraints. The integration of renewable energy sources, like wind and solar, however, poses challenges due to their inherent variability. This variability, driven largely by changing weather conditions, demands frequent recalibrations of power settings, thus necessitating recurrent OPF resolutions. This task is daunting using traditional numerical methods, particularly for extensive power systems. In this work, we present a cutting-edge, physics-informed machine learning methodology, trained using imitation learning and historical European weather datasets. Our approach directly correlates electricity demand and weather patterns with power dispatch and generation, circumventing the iterative requirements of traditional OPF solvers. This offers a more expedient solution apt for real-time applications. Rigorous evaluations on aggregated European power systems validate our method's superiority over existing data-driven techniques in OPF solving. By presenting a quick, robust, and efficient solution, this research sets a new standard in real-time OPF resolution, paving the way for more resilient power systems in the era of renewable energy.
\end{abstract}

% \includegraphics{figs/grabs.pdf}

% \begin{highlights}
% \item Research highlights item 1
% \item Research highlights item 2
% \item Research highlights item 3
% \end{highlights}

% Keywords

\maketitle

\nomenclature[A]{OPF}{Optimal power flow}
\nomenclature[A]{ACOPF}{Alternating current optimal power flow}
\nomenclature[A]{DCOPF}{direct current optimal power flow}
\nomenclature[A]{LMP}{Locational marginal price}
\nomenclature[A]{RES}{Renewable energy sources}
\nomenclature[A]{PV}{Photovoltaic}
\nomenclature[A]{AI}{Artificial intelligence}
\nomenclature[A]{ML}{Machine learning}
\nomenclature[A]{DT}{Decision tree}
\nomenclature[A]{CVAE}{Conditional variational autoencoder}
\nomenclature[A]{GAT}{Graph attention network}
\nomenclature[A]{RNN}{Recurrrent neural network}
\nomenclature[A]{CNN}{Convolutional neural network}
\nomenclature[A]{NNs}{Neural networks}
\nomenclature[A]{PCA}{Principal component analysis}
\nomenclature[A]{GCN}{Graph convolutional network}
\nomenclature[A]{DNN}{Deep neural network}
\nomenclature[A]{SMW-GSAT}{Spacial multi-window graph self-attention neural network}
\nomenclature[A]{NLAT}{Node-link attention network}
\nomenclature[A]{MLP}{Multilayer perceptron}
\nomenclature[A]{LeakyReLU}{Leaky version of a Rectified Linear Unit}
\nomenclature[A]{PyPSA}{Python for power system analysis}
\nomenclature[A]{OCGT}{open cycle gas turbine}
\nomenclature[A]{KNN}{K-nearest neighbors}
\nomenclature[A]{API}{Application programming interface}
\nomenclature[A]{MAAPE}{Mean arctangent absolute percentage error}
\nomenclature[A]{PCs}{Principal components}
\nomenclature[A]{LR}{linear regressor}
\nomenclature[A]{SVR}{Support vector regressor}
\nomenclature[A]{IP}{interior point}
\nomenclature[N]{$P_{i}^{\textrm{G}}$}{Active power output of generator at node $i$}
\nomenclature[N]{$Q_{i}^{\textrm{G}}$}{Reactive power output of generator at node $i$}
\nomenclature[N]{$P_{i}^{\textrm{D}}$}{Active power consumption at node $i$}
\nomenclature[N]{$Q_{i}^{\textrm{D}}$}{Reactive power consumption at node $i$}
\nomenclature[N]{$G_{jk}$}{Conductance between node $j$ and $k$}
\nomenclature[N]{$B_{jk}$}{Susceptance between node $j$ and $k$}
\nomenclature[N]{$V_{j}$}{Voltage magnitude at node $j$}
\nomenclature[N]{$\theta _{jk}$}{Voltage angle difference between node $j$ and $k$}
\nomenclature[N]{$F_{jk}$}{Active power flow along the transmission line joining node $j$ and $k$}
\nomenclature[N]{$\mathscr{G}$}{Graph data structure}
\nomenclature[N]{$\mathcal{N}$}{Set of nodes in the graph}
\nomenclature[N]{$\mathcal{L}$}{Set of edges in the graph}
\nomenclature[N]{$\mathcal{N}_{\textrm{G}}$}{Subset which contains nodes that have controllable generators}
\nomenclature[N]{$\mathcal{N}^{i}$}{Subset which contains nodes connected to node $i$}
\nomenclature[N]{$\bar{P}_{i}^{\textrm{G}}, \bar{Q}_{i}^{\textrm{G}}, \bar{V}_{j}$ , $\bar{\theta }_{jk}$}{Upper bounds}
\nomenclature[N]{$\underaccent{\bar}{P}_{i}^{\textrm{G}}, \underaccent{\bar}{Q}_{i}^{\textrm{G}}, \underaccent{\bar}{V}_{j}$ , $\underaccent{\bar}{\theta }_{jk}$}{Lower bounds}
\nomenclature[N]{$\bar{F}_{jk}$}{Nominal power on transmission line connect node $j$ and $k$}
\nomenclature[N]{$\mathfrak{F}$}{Overall mapping function}
\nomenclature[N]{$\boldsymbol{\lambda}$}{Parameters in the proposed neural network model}
\nomenclature[N]{$\mathbf{P}^{\textrm{D}}$}{Power demand tensor}
\nomenclature[N]{$\boldsymbol{\eta }$}{Weather condition tensor}
\nomenclature[N]{$\hat{\mathbf{P}}^{\textrm{G}}$}{Optimal generator active output tensor}
\nomenclature[N]{$\hat{\mathbf{F}}$}{Optimal active power dispatch}
\nomenclature[N]{$\mathbf{S}$}{Feature matrix of the graph}
\nomenclature[N]{$N$}{Number of nodes}
% \nomenclature[N]{$d$}{Length of node feature}
\nomenclature[N]{$\mathbf{A}$}{Adjacency matrix of the graph}
\nomenclature[N]{$\mathbf{I}$}{Identity matrix}
\nomenclature[N]{$\mathbf{D}$}{Degree matrix of the graph}
\nomenclature[N]{$\mathbf{Q}$}{Eigenvectors}
\nomenclature[N]{$\mathbf{L}$}{Laplacian matrix}
\nomenclature[N]{$\mathbf{P}_{\textrm{node}}$}{Positional encoding for the nodes}
% \nomenclature[N]{$m$}{Length of node positional encoding}
\nomenclature[N]{$F$}{Length of input node feature of proposed model}
\nomenclature[N]{$\mathbf{H}$}{Input matrix of proposed model}
\nomenclature[N]{$\mathfrak{E}$}{SMW-GSAT layer}
\nomenclature[N]{$\mathbf{W}$}{Transformation matrices in SMW-GSAT layer}
\nomenclature[N]{$\mathbf{a}$}{Attention vector in SMW-GSAT layer}
\nomenclature[N]{$\mathbf{H}'$}{Node feature matrix after graph attention}
\nomenclature[N]{$F'$}{Length of node feature in matrix $\mathbf{H}'$}
\nomenclature[N]{$K$}{Number of windows in multi-window mechanism}
\nomenclature[N]{$\mathbf{H}''$}{Node feature matrix after SMW-GSAT layer}
\nomenclature[N]{$\mathbf{P}_{\textrm{link}}$}{}
\nomenclature[N]{$L$}{Number of links}
\nomenclature[N]{$\mathfrak{G}$}{NLAT layer}
\nomenclature[N]{$\mathbf{W}_{\textrm{Q}}, \mathbf{W}_{\textrm{K}},\mathbf{W}_{\textrm{V}}$}{Transformation matrices in NLAT layer}
\nomenclature[N]{$V$}{Length of last dimension of quires and keys matrices}
\nomenclature[N]{$U$}{Number of latent features for each link}
\nomenclature[N]{$\mathfrak{H}$}{MLP layer}
\nomenclature[N]{$\mathbf{W}',\mathbf{b}$}{Weight matrix and bias vector in MLP layer}
\nomenclature[N]{$R$}{Number of hidden layers in MLP}

\section{Introduction}

The simplest form of Optimal power flow (OPF) is to determine the optimal power production from each generators within the power grid in order to meet the demand of electricity consumption, meanwhile satisfying physical and engineering constraints. As a crucial aspect in energy management, OPF has been defined since 1962 and has many variants in the process of development according to different formulations and constraints it contains \cite{cain2012history}. 

One of the variants that use exact alternating current formulation is known as ACOPF. In addition to determining the active and reactive power output from generators, other control variables in the power grid such as voltage magnitude and voltage angle are also determined subject to their constraints. Due to the sinusoidal nature of alternating current, the optimization problem becomes nonlinear and non-convex. Consequently, ACOPF has been demonstrated to be an NP-hard problem\cite{bienstock2019strong}, making it not only costly to solve but also challenging to achieve its global optimum.

To mitigate this challenge, linear approximations can be employed, effectively linearizing the AC system into a direct current (DC) format. By making certain approximations regarding voltage magnitude and angle, DCOPF eliminates the sinusoidal formulation, streamlining the problem and decreasing computational complexity. Although these approximations might result in solutions that are somewhat sub-optimal and less reliable, such solutions remain valuable in the transmission system. This is especially true since voltages are typically maintained within a tight range close to nominal values, and reactive power isn't central to the primary system analyses. In industrial settings, many software tools consistently employ DCOPF for simulating, analyzing, and forecasting Locational Marginal Price (LMP) \cite{sharma2016comparative}

However, with the increasing penetration of renewable energy sources (RES), such as wind and solar power generators, solving the OPF problem becomes more significant and frequent. The uncertain nature of large-scale integration of variable RES makes it technically challenging to keep the power system flexible \cite{mladenov2021impact,mladenov2018characterisation}. Flexibility maintenance in power systems requires providing supply-demand balance, maintaining continuity in unexpected situations, and coping with uncertainty on supply-demand sides \cite{impram2020challenges}, which is one of the main objects in OPF problem. Solar power is determined by solar irradiation and wind power is determined by the wind speed, i.e., the meteorological condition, which changes in very short time intervals especially for wind. Since different RES supply situations lead to different power grid operation settings, the OPF problem needs to be solved in several minutes' period or even less, i.e., in real-time \cite{tong2011look}, in order to serve the power demand while ensuring stable power grid operations. 

% \begin{nomenclature}
% \begin{deflist}[A] 
% \defitem{A}\defterm{radius of}
% \defitem{B}\defterm{position of}
% \defitem{C}\defterm{further nomenclature continues down the page}
% \end{deflist}
% \end{nomenclature}\begin{table*}[!t]
\begin{table*}[!t]
  \begin{framed}
    \printnomenclature[2.5cm]
  \end{framed}
\end{table*}

Traditionally, the OPF problem has been approached as a classical optimization challenge, 
often modeled using methods like linear programming \cite{low2014convex} or quadratic programming \cite{momoh1989generalized}. In earlier days, 
it was solved using conventional optimization techniques such as the Newton method \cite{rashed1974optimal}, simplex method \cite{wells1968method}, 
or the interior point method \cite{momoh1999review}. These methods have seen continuous improvements over time \cite{momoh1999review1, momoh1999review2}. 
However, when applied to larger-scale power systems, these iteration-based methods significantly ramp up computational complexity and time consumption \cite{capitanescu2011state}. 
This escalation complicates real-time management of power systems with a high reliance on renewable sources.

With the advancements of computational capability and artificial intelligence (AI) techniques, there are also many attempts to use modern AI methods to solve OPF problems in model-less ways. One line of approach is using heuristic algorithms, e.g., genetic algorithm \cite{bakirtzis2002optimal}, particle swarm optimization \cite{park2009improved}, artificial bee colony algorithm \cite{adaryani2013artificial}, and hybrid algorithms \cite{ks2019hybrid, hassanien2018hybrid} in order to overcome drawbacks when using single algorithm. They have the advantage of efficiently handling nonlinear and non-convex problems compared to other traditional numerical methods. However, there are still two main limitations: the first comes from the lack of the guarantee of optimality and feasibility; and the second lies in the high computational demand needed, which still prevents these methods from a real-time scenario deployment \cite{vikhar2016evolutionary}. Another line of approach is using machine learning (ML) methods, which hold the promise to address these limitations. Since most of the ML algorithms can be trained offline and deployed online in realistic scenarios, they can be used to provide acceptable real-time solutions efficiently. Various ML algorithms have already been successfully applied to many kinds of problems in power systems, e.g., using decision tree (DT) to solve generator strategic bidding problem \cite{prat2022learning}, using conditional variational autoencoder (CVAE) to generate synthetic load profiles \cite{wang2022generating}, using graph attention network (GAT) to forecast multi-site photovoltaic (PV) power \cite{simeunovic2022interpretable}, using recurrent neural network (RNN) or convolutional neural network (CNN) to perform instability or fault assessment \cite{veerasamy2021lstm, wen2019real,shi2020convolutional}. See \cite{mosavi2019state} for a comprehensive survey of ML applications in power systems. 

While using ML to deal with OPF problems, research can be divided into two categories. Some works focus on using ML to assist conventional OPF solvers by reducing the complexity of the original optimization problem, e.g., using neural networks (NNs) to represent system security boundary, from which a differentiable mapping function can be extracted, and integrated then into the original OPF model \cite{gutierrez2010neural}; using principal components analysis (PCA) to map OPF equations to a new domain, reducing the dimensionality of the OPF problem \cite{vaccaro2016knowledge}; using NNs to predict a set of active constraints at optimality, thus reducing the size of feasible solution space which is going to be searched \cite{deka2019learning, guha2019machine}. Other research aim at using ML to build up an end-to-end approach, directly predicting solutions for OPF problem, e.g., combining feed-forward network (FFN) and Lagrangian dual method, predicting the OPF solution while ensuring the satisfaction of physical and engineering constraints \cite{fioretto2020predicting}; or using graph convolutional network (GCN) and imitation learning to compute OPF solutions \cite{owerko2020optimal}, which leverages the topological structure of power grid by introducing the adjacency matrix; using deep neural network (DNN) to predict OPF solution while ensuring the feasibility by constraints calibration \cite{zhao2020deepopf+} or post-processing \cite{pan2020deepopf}. A hybrid method developed not only a data-driven QPF regression model but also incorporated a sample pre-classification strategy based on active constraints identification \cite{lei2020data}. 

However, there are still drawbacks of these ML-based approaches mentioned above. For those approaches that use ML to assist conventional solvers, only part of the optimization process is replaced by ML model or part of the optimization problem is simplified, iterations for optimality searching are still needed, thus time complexity can be high especially considering a large-scale power system. For those approaches using ML to directly generate OPF solutions, a way to ensure a feasible solution becomes the main focus, nevertheless, some approaches can only provide soft restrictions such as adding Lagrangian terms into the loss function. Although post-processing, which projects an infeasible solution onto the surface of a feasible domain, is able to avoid this problem and yields an absolutely feasible solution, all these approaches lack interpretability.

In this paper, we solve DCOPF problem by using a proposed physics-informed neural network model including a spacial multi-window graph self-attention network (SMW-GSAT), a node-link attention network (NLAT) under imitation learning framework, directly mapping the power demand and weather condition to power dispatches and generators output settings. More specifically, we introduced SMW-GSAT layer to encode the node features in high-dimensional feature space, and masked graph self-attentions were executed in parallel in each window inside SMW-GSAT in order to integrate information in different patterns and increase the expressive power of network. We then considered the encoded node features as the context of the state of the nodes, and applied NLAT layer to capture the correlation between active power transmission links and nodes, accordingly converted node state contexts into latent features for transmission links. At last, we applied multilayer perceptron (MLP) to further decode the state of transmission links, yielding the power dispatched on each link. Post-processing was applied to ensure the feasibility. The proposed machine learning framework was trained to imitate the optimization process of an interior point solver Gurobi by using optimal solutions obtained from it, whose interfaces are integrated in Python for Power System Analysis (PyPSA) toolbox. The contributions of this paper can be summarized as follows:
\begin{itemize}
    \item A physics-informed neural network model is proposed including SMW-GSAT and NLAT for solving optimal power flow problems. They can be used to promptly generate feasible sub-optimal solutions considering fluctuating wind and solar resources.
    \item The layout of power grid is integrated into the multi-window mechanism, pushing the model to learn different attention matrices which indicate the different range of affections mainly determined by the weather.
    \item The involved neural network is interpreted to some extent by looking into the attention matrices in corresponding test cases. 
    \item The performance of the proposed physics-informed neural network model is evaluated on the test set and compared to conventional ML algorithms and other data-driven approaches.
\end{itemize}
The remainder of this paper is organized as follows. Section \ref{Sec2} introduces preliminaries on the definition of the OPF problem. Section \ref{Sec3} details the proposed imitation learning framework, and the definition of neural network layers inside the structure. Section \ref{Sec4} describes the experiments conducted by this work after introducing the data flow. Results and discussions of case studies are exhibited in section \ref{Sec5}. At last, conclusions are drawn in section \ref{Sec6}.

\section{Formulation of OPF Problem} \label{Sec2}

In this section, we introduce the mathematical representations of the ACOPF and DCOPF problems. 
We can model the entire power network as a graph $\mathscr{G}(\mathcal{N}, \mathcal{L} )$. The node set $\mathcal{N}$ consists of $N$ buses, and the edge set $\mathcal{L}$ consists of $L$ transmission lines. The subset $\mathcal{N}_{\textrm{G}}$ within $\mathcal{N}$ denotes nodes that have controllable generators. The subset $\mathcal{N}^{i}$ within $\mathcal{N}$ represents nodes connected to node $i$. The ACOPF determines the cost-optimal generator active outputs that meet the power demand over the power grid, subject to physical and engineering constraints. The ACOPF can be expressed as:
\begin{equation}
    \underset{P_{i}^{\textrm{G}}} {\textrm{minimize}}\quad \sum_{i\in\mathcal{N}_{\textrm{G}}} \mathrm{cost}(P_{i}^{\textrm{G}})
    \label{eq_2_cost_func}
\end{equation}
subject to
\begin{equation}
\begin{aligned}
    &P_{j}^{\textrm{G}}-P_{j}^{\textrm{D}}=V_{j}\sum _{k\in \mathcal{N}^{j}}V_{k}(G_{jk}\textrm{cos}\theta _{jk}+B_{jk}\textrm{sin}\theta _{jk})\\
    &Q_{j}^{\textrm{G}}-Q_{j}^{\textrm{D}}=V_{j}\sum _{k\in \mathcal{N}^{j}}V_{k}(G_{jk}\textrm{sin}\theta _{jk}-B_{jk}\textrm{cos}\theta _{jk})
    %\quad j\in \textbf{N}
    \label{eq_2_node_bal}
\end{aligned}
\end{equation}
\vspace{-15pt}
\begin{equation}
    \underaccent{\bar}{P}_{i}^{\textrm{G}} \leq P_{i}^{\textrm{G}} \leq \bar{P}_{i}^{\textrm{G}}
    \label{eq_2_acpower_lim}
\end{equation}
\vspace{-15pt}
\begin{equation}
    \underaccent{\bar}{Q}_{i}^{\textrm{G}} \leq Q_{i}^{\textrm{G}} \leq \bar{Q}_{i}^{\textrm{G}}
    \label{eq_2_reacpower_lim}
\end{equation}
\vspace{-15pt}
\begin{equation}
    \underaccent{\bar}{V}_{j} \leq V_{j} \leq \bar{V}_{j} 
    \label{eq_2_voltagemag_lim}
\end{equation}
\vspace{-15pt}
\begin{equation}
    \underaccent{\bar}{\theta }_{jk} \leq \theta _{jk} \leq \bar{\theta }_{jk}
    \label{eq_2_voltageang_lim}
\end{equation}

where $i\in\mathcal{N}_{\textrm{G}}$, $j\in\mathcal{N}$, $k\in \mathcal{N}^{j}$, $P_{j}^{\textrm{G}}$ and $ Q_{j}^{\textrm{G}}$ are setting points that represent the active and reactive power output of at node $i$, $P_{j}^{\textrm{D}}$ and $Q_{j}^{\textrm{D}}$ represent the active and reactive power consumption at node $j$, $V_{j}$ and $\theta _{jk}$ represent the voltage magnitude at node $j$ and voltage angle difference between node $j$ and $k$. $G_{jk}$ and $B_{jk}$ represent the conductance and susceptance between node $j$ and $k$. $\bar{P}_{i}^{\textrm{G}}, \bar{Q}_{i}^{\textrm{G}}, \bar{V}_{j}$ and $\bar{\theta }_{jk}$ are upper bounds of each variables. $\underaccent{\bar}{P}_{i}^{\textrm{G}}, \underaccent{\bar}{Q}_{i}^{\textrm{G}}, \underaccent{\bar}{V}_{j}$ and $\underaccent{\bar}{\theta }_{jk}$ are lower bounds of each variables. 

The objective function, as defined in \eqref{eq_2_cost_func}, calculates the cumulative cost associated with the operation of the power system. This cost is primarily determined by the active generator output, represented by the function $\mathrm{cost}()$. Generally, this function assumes a quadratic form. The equality constraints in formulation \eqref{eq_2_node_bal}, commonly referred to as the power flow equations or nodal power balance constraints, stem from fundamental electrical principles: Ohm's Law and Kirchhoff's Current Law \cite{cain2012history}. The constraints expressed in \eqref{eq_2_acpower_lim} and \eqref{eq_2_reacpower_lim} set the boundaries for active and reactive power outputs, respectively. Similarly, \eqref{eq_2_voltagemag_lim} and \eqref{eq_2_voltageang_lim} define the permissible limits for voltage magnitude and the difference in voltage angles, respectively.

The DCOPF model offers a linear approximation of the ACOPF problem, as elucidated in \cite{wood2013power}. The transition from ACOPF to its DCOPF counterpart rests on three fundamental assumptions:

\begin{enumerate}[1)]
    \item Every transmission line's resistance is overlooked, making the term $G_{jk}$ zero.
    \item The voltage magnitude at each node is consistently set to the nominal value, i.e., 1.
    \item Voltage angle differences across node pairs are assumed to be diminutive, generally less than $\pi/6$, which leads to the approximations $\textrm{cos}\theta _{jk} \approx 1$ and $\textrm{sin}\theta _{jk} \approx \theta _{jk}$.
\end{enumerate}

With these premises, the reactive component of the nodal balance constraint, represented by \eqref{eq_2_node_bal}, becomes redundant. The active component subsequently reformulates as:
\begin{equation}
    P_{j}^{G}-P_{j}^{D}=\sum _{k\in \mathcal{N}^{j}}B_{jk}\theta _{jk}=\sum _{k\in \mathcal{N}^{j}}F_{jk}
\end{equation}
Herein, $F_{jk}$ denotes the active power flow along the transmission line joining nodes $j$ and $k$ \cite{horsch2018linear}. Correspondingly, the inequality constraints given by \eqref{eq_2_voltagemag_lim} and \eqref{eq_2_voltageang_lim} transform into:
\begin{equation}
    -\bar{F}_{jk} \leq F_{jk}\leq \bar{F}_{jk}\qquad \forall k\in \mathcal{N}^{j}
    \label{eq_2_act_flow_lim}
\end{equation}
In this context, $\bar{F}_{jk}$ denotes the nominal power traversing the transmission line between node $i$ and its adjacent node $k$. A positive $F_{jk}$ denotes power withdrawal from node $i$, whereas a negative value implies power withdrawal from node $k$.

\section{Attention based network for OPF Solving} \label{Sec3}

\subsection{The Overall Architecture}
\begin{figure*}[h]
    \begin{center}
        \includegraphics[width=0.7\textwidth]{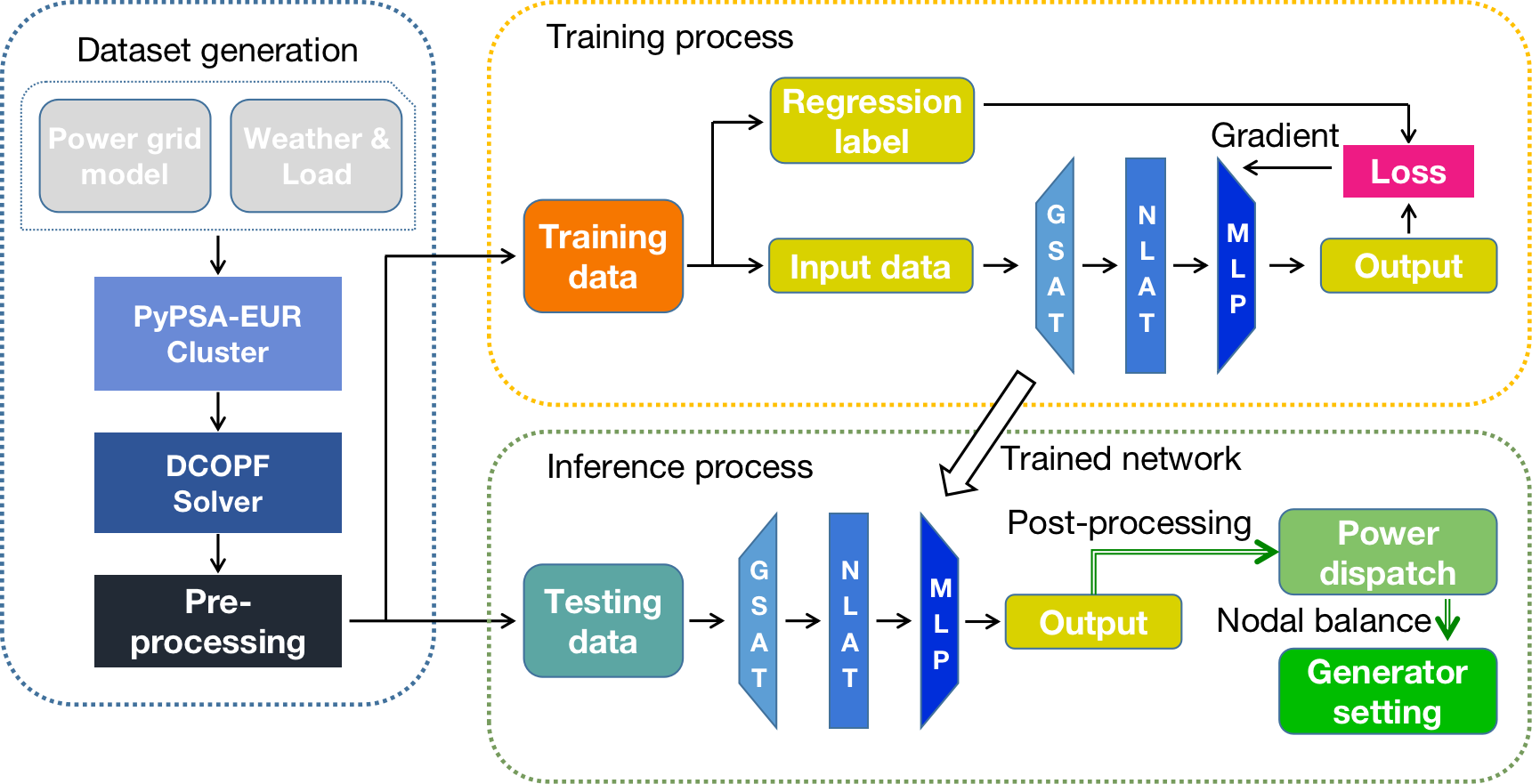}
    \end{center}
    \caption{The flowchart of proposed ML framework for OPF.}
    \label{fig:flowchart}
\end{figure*}

We propose a ML based framework for the OPF problem in a highly renewable power system, a flow chart is shown in Fig. \ref{fig:flowchart}. In the framework we focused on the design of ML architecture, which consisted three different layers performing different functions.

We introduced SMW-GSAT layer to encode the node features in high-dimensional feature space. SMW-GSAT consists of graph self attention and multi-window mechanism. Graph attention is responsible for capturing the correlations between different nodes in the power system, different from GCN graph attention is able to vary aggregation weights based on the input weather conditions, forcing nodes focus on part of the neighbours. This enhanced the flexibility of network by identifying and focusing on the most influential nodes in the graph representing the power system. Masked multi-window mechanism allows graph attention integrate information in different patterns as well as being executed in parallel in each window, increasing the expressive power of network. Without the SMW-GSAT, the model might miss these localized spacial correlations and interpretability, potentially leading to less accurate or less efficient solutions. We then considered the encoded node features as the context of the state of the nodes, and applied NLAT layer to capture the correlation between active power transmission links and nodes, accordingly converted node state contexts into latent features for transmission links. One of the advantages of NLAT is it captures the interactions between components, which helps the model account for the inherent interconnectivity of power systems, where changes in one component can have cascading effects on system outputs. Another advantage of NLAT is similar to SMW-GSAT, it captures localized correlations. Without the NLAT, the model might overlook the complex dependencies that exist within the power grid. At last, we applied MLP to further decode the state of transmission links, yielding the power dispatched on each link. The inclusion of an MLP in the architecture mitigates the task of NLAT layer and further enhances the nonlinearity, improving the model's ability to predict the optimal power dispatch in each link. Without the MLP, the model might be closer to linear assumptions, potentially reducing its ability to accurately solve the OPF problem in real-world power systems, which often exhibit nonlinear behavior.

Looking at the whole framework it contains three main phases, which are data generation, training and inference. The data generation and training phases are done offline since they are time-consuming considering a large dataset. The inference phase is used online to solve OPF problems in real-time scenarios.

We treated the neural network model as a mapping function, indicating the relationship between optimal power dispatches and input power demand, and weather data:
\begin{equation}
    \mathfrak{F}_{\boldsymbol{\lambda}}\left ( \left\{ \mathbf{P}^{\textrm{D}},\:\boldsymbol{\eta }\right\} \right )=\left\{ \hat{\mathbf{P}}^{\textrm{G}}, \: \hat{\mathbf{F}}\right\}
\label{eq_3_projection}
\end{equation}
where $\mathfrak{F}$ is the parameterized mapping function, $\boldsymbol{\lambda}$ denotes the parameters in the proposed neural network model, $\mathbf{P}^{\textrm{D}}$ and $\boldsymbol{\eta }$ are input features correspond to the tensor of power demand and tensor of weather condition respectively, $\hat{\mathbf{P}}^{\textrm{G}}$ and $ \hat{\mathbf{F}}$ are optimal outputs correspond to the tensor of active generator power and tensor of active power dispatch respectively. 

During the dataset generation phase, we initially gathered parameters of the power system and weather data from an open-source dataset. Following this, we conducted a clustering of the original power grid down to the desired size. Once this was accomplished, we computed the Optimal Power Flow (OPF) solutions for the clustered power grid. These solutions were then split into two distinct sets, one for training purposes and the other for testing. 

During the training phase, we trained our proposed neural network model to perform the regression of active power dispatch across each transmission line in the power grid. The proposed model integrates two attention-based structures: the Spatial Multi-Window Graph Self Attention Layer (SMW-GSAT) and the Node-Link Attention Layer (NLAT). And a Multilayer Perceptron (MLP) as the final stage. 

During the inference phase, we utilized the trained neural network to produce the optimal active power dispatch. Subsequently, we implemented post-processing measures to guarantee the feasibility of the solution. Finally, we computed the active power output for each generator, adhering to the power balance constraints at each node. Detailed information on the proposed framework will be elaborated in the subsequent subsections.

\subsection{Input Data}
We begin by detailing the input data structure for the neural networks. The power grid's state at a given time is represented as graph data, comprising a feature matrix $\mathbf{S} = \{ \mathbf{s}_i | i \in \mathcal{N} \} \in \mathbb{R}^{d \times N}$ and an adjacency matrix $\mathbf{A} \in \mathbb{R}^{N \times N}$. Here, $\mathcal{N}$ represents the set of nodes in the graph, with $N$ being the total node count. Each node is characterized by $d$ features, such as power consumption and weather conditions. The column vector $\mathbf{s}_i \in \mathbb{R}^d$ represents node $i$'s features. The adjacency matrix $\mathbf{A}$ depicts the power grid's layout: if nodes $i$ and $j$ are adjacent, the element $a_{i,j}$ is 1; otherwise, it's 0.

In the OPF problem, variations in independent input variables can elicit starkly different power system responses, especially with respect to weather input. To enhance the neural networks' ability to recognize and adapt to this
%To bolster the capacity of neural networks to capture 
this non-translation invariance, we integrated positional information into the node features using Laplacian positional encoding (LPE) \cite{dwivedi2020generalization,belkin2003laplacian}. The primary intent of LPE is straightforward: nodes in proximity should have analogous positional features, while distant nodes should possess distinct features. These encodings are obtained through the eigendecomposition of the normalized Laplacian matrix $\mathbf{L}$:
\begin{equation}
    \mathbf{L} = \mathbf{I} - \mathbf{D}^{-\frac{1}{2}}\mathbf{A}\mathbf{D}^{-\frac{1}{2}} = \mathbf{Q}\mathbf{\Lambda} \mathbf{Q}^{-1}
\end{equation}
Here, $\mathbf{I}$ is the identity matrix, $\mathbf{D} \in \mathbb{R}^{N\times N}$ is the degree matrix, $\mathbf{Q}\in \mathbb{R}^{N\times N}$ contains the eigenvectors of $\mathbf{L}$ as columns, and $\mathbf{\Lambda} \in \mathbb{R}^{N\times N}$ is a diagonal matrix with the corresponding eigenvalues on the diagonal. 
We selected the first $m$ smallest non-trivial eigenvectors $\mathbf{P}_{\textrm{node}}\in\mathbb{R}^{N\times m}$ as the positional encoding for the nodes, represented as $\mathbf{P}_{\textrm{node}}^{\intercal}=\{\mathbf{p}_{\textrm{node},i}|i\in \mathcal{N}\}\in \mathbb{R}^{m\times N}$. These LPEs are then appended to the feature matrix, creating the following input data:

\begin{equation}
    \mathbf{H}=\mathbf{S}\|\mathbf{P}_{\textrm{node}}^{\intercal}=\{\mathbf{h}_i|i\in \mathcal{N}\}\in \mathbb{R}^{F\times N}
\end{equation}
where $\intercal$ signifies matrix transposition, $\|$ indicates concatenation, and $F=d+m$ represents the total number of input features per node. 
The column vector $\mathbf{h}_i \in \mathbb{R}^{F}$ encapsulates the entire feature set for node $i$.

\subsection{Spacial Multi-window Graph Self Attention Layer}
We used a graph attention network (GAT) to capture the interrelations between nodes, anticipating that these correlations would change based on the weather input. The SMW-GSAT is an adaptation of a multi-head graph attention network \cite{velivckovic2017graph}.

The SMW-GSAT layer can be viewed as a function mapping from the original feature space to a higher-dimensional feature space : 
$\mathfrak{E}_{\mathbf{W}, \mathbf{a}}\,:\,\mathbb{R}^{F\times N}\,\to\,\mathbb{R}^{F'\times N}$, where $\mathfrak{E}_{\mathbf{W}, \mathbf{a}}$ denotes the SMW-GSAT layer parameterized by transformation matrices $\mathbf{W} \in \mathbb{R}^{N\times F\times F{'}}$ and attention vector $\mathbf{a}\in \mathbb{R}^{2F{'}}$. Note that the transformation matrices are different for each node, in order to augment express power. We therefore got a new feature matrix $\mathbf{H}'=\left\{ \mathbf{h}'_{i}|i\in \mathcal{N}\right\}\in \mathbb{R}^{F'\times N}$ in higher dimension, where column vector $\mathbf{h}'_{i}\in\mathbb{R}^{F'}$ denotes output features for node $i$.

\begin{figure}[t]
    \centering
    \begin{subfigure}{1.\linewidth}
        \centering
        \includegraphics[width=\linewidth]{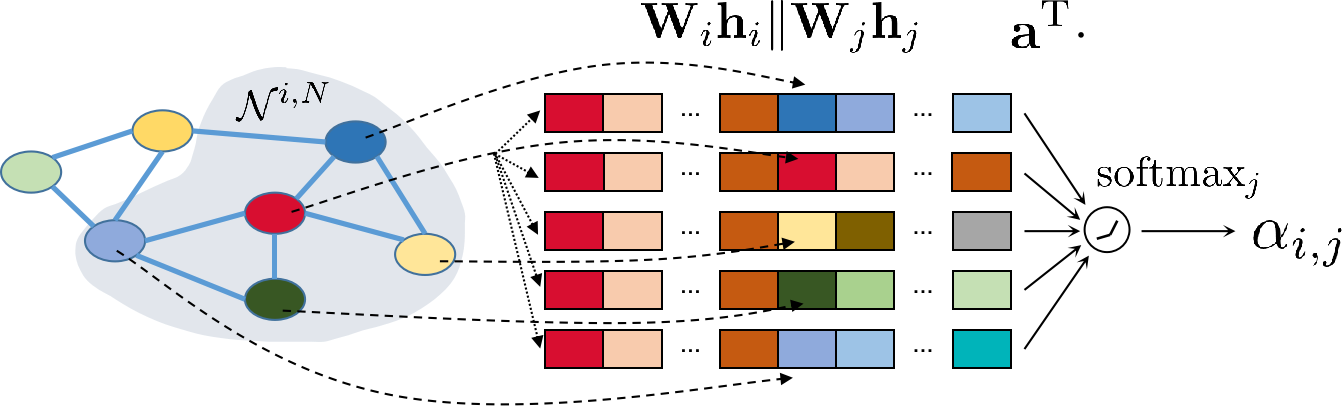}
        \caption{}
        \label{fig:GAT_1}
    \end{subfigure}
    \par\medskip
    \begin{subfigure}{.6\linewidth}
        \centering
        \includegraphics[width=\linewidth]{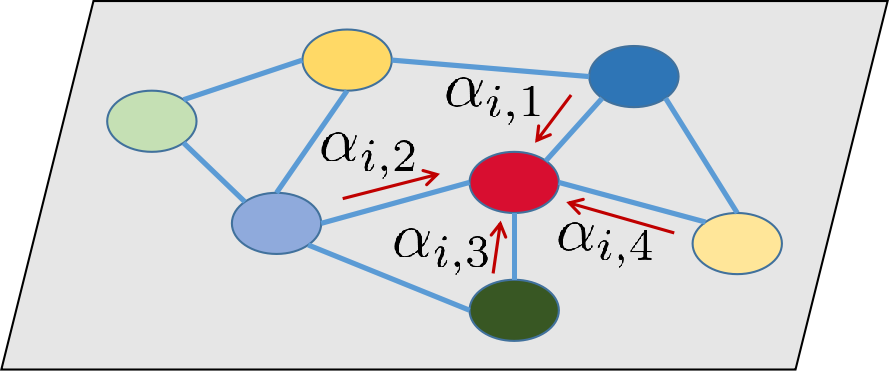}    
        \caption{}
        \label{fig:GAT_2}
    \end{subfigure} 
    \caption{Masked spacial graph self-attention mechanism.}
    \label{fig:GAT}
\end{figure}

Then the correlation between nodes $i$ and $j$ can be calculated as follows:
\begin{equation}
    e_{i,j} =  \mathrm{LeakyReLU\left ( \mathbf{a}^{\intercal}\cdot \left [ \mathbf{W}_{i}^{\intercal}\mathbf{h}_{i}\middle\| \mathbf{W}_{j}^{\intercal}\mathbf{h}_{j}\right ] \right )}
\end{equation}
\vspace{-15pt}
\begin{equation}
    \alpha _{i,j} = \mathrm{softmax}_{j}\left ( e_{i,j} \right ) = \frac{\mathrm{exp}(e_{i,j})}{\sum_{k}\mathrm{exp}(e_{i,k})}
\end{equation}
where $e_{i,j}$ denotes the self-attention coefficient of node $i$ while focusing on its neighbor node $j$, and $\mathbf{W}_{i},\mathbf{W}_{j} \in \mathbb{R}^{F\times F{'}}$ denote the linear transformation matrix for node $i$ and $j$, respectively. They two are different in order to increase the express ability of NNs. The dot product $\cdot$ with $\mathrm{LeakyReLU}$ activation function was applied to generate the correlation. We applied $\mathrm{softmax}$ to get the normalized attention scores $\alpha_{i,j}$ so that they can be compared more easily. Figure.~\ref{fig:GAT_1} illustrates the calculation process of attention scores.

In common cases this correlation is calculated between each node and all other nodes in the graph. However, we integrated the layout information into this calculation by using masked attention, which makes the node focus on a specific subset of nodes. We calculated the attention score only between node $i$ and its neighboring nodes $j \in \mathcal{N}^{i,T}$ including node $i$ itself, where $\mathcal{N}^{i,T}$ denotes the set of neighborhoods inside the range of $T$ jumps for node $i$. We defined the support of neighborhoods by using the concept of $t$-hop neighborhoods, which measures the distance of two nodes according to the structure of the graph rather than the geographical location. Nodes that are indirectly connected with node $i$ via a shortest path including $t$ links and $t-1$ nodes are $t$-hop neighborhoods of node $i$. Set of neighborhoods $\mathcal{N}^{i,T}$ includes all $t$-hop neighborhoods where $t=0,1,...,T$, it can be derived by $N$-th power as follows:
\begin{equation}
    \mathcal{N}^{i,T} = \textbf{where}\left ( \left ( \mathbf{A} + \mathbf{I}\right )^{T}_{i}>0 \right )
\end{equation}
where $\mathbf{A}$ denotes the adjacency matrix of graph, $\mathbf{I}$ denotes the identity matrix,  $()_{i}$ denotes the $i$-th row of a matrix, $\textbf{where}()$ denotes the function giving the position of true input. If no node is isolated, i.e., the graph is connected, the limitation of $n$-hop neighborhoods covers all the nodes in the graph. Then we conduct feature aggregation by using $\alpha_{i,j}$:

\begin{equation}
    \mathbf{h}'_{i}=\underset{j\in\mathcal{N}^{i,T}}{\sum}\alpha_{i,j}\mathbf{W}_{j}^{\intercal}\mathbf{h}_{j}
    \label{eq_3A_att}
\end{equation}

Fig.~\ref{fig:GAT_2} depicts the feature aggregation. Furthermore, graph attention can be extended to multi-head graph attention, so that the learning process can be stabilized, it can also provide diverse attention areas \cite{huang2021topology}. Similar to multi-head graph attention, we introduced a multi-window graph attention mechanism instead. Different from \cite{simeunovic2022interpretable}, we used different ranges of neighborhoods as the mask in each attention window, to force different windows to focus on different ranges of neighborhoods. Specifically, each window conducted node feature embedding through \eqref{eq_3A_att} in parallel, then we concatenated all the embedded features from all $K$ independent windows:
\begin{equation}
    \mathbf{h}_{i}''=\underset{k=1}{\overset{K}{\|}}\underset{j\in\mathcal{N}_{k}^{i,T_{k}}}{\sum}\alpha_{i,j}^{k}(\mathbf{W}_{j}^{k})^{\intercal}\mathbf{h}_{j}
\end{equation}
where $\alpha_{i,j}^{k}$ denote normalized attention scores computed in $k$-th window, and the attention mechanism in $k$-th window is parameterized by $\mathbf{W}_{j}^{k}$ and $\mathbf{a}^{k}$, $\mathcal{N}_{k}^{i,T_{k}}$ denotes the set of neighborhoods of node $i$ within range of $T_{k}$ in $k$-th window, and $\|$ denotes concatenation for embedded features coming from each window. Note that, the output high-dimensional feature matrix of SMW-GSA layer is $\mathbf{H}''\in  \mathbb{R}^{KF{'}\times N}$, we treated it as the context of the node states.

Thus, we used SMW-GSA layer to capture the dynamic correlations that should be different in each spatial attention window, and the correlation matrix underscores the significance of nodes according to different states of the power grid. It provides us an opportunity to interpret the network while addressing the OPF challenge.

\subsection{Node Link Attention Layer}
Inspired by transformer \cite{vaswani2017attention}, we proposed a decoder-like layer using an attention mechanism, NLAT converted the input context of the node states to an expression of links. 
With the fact that the outputs of this layer, i.e., link expressions on a certain graph, have no order compared to the words in a sentence from Natural language processing (NLP), therefore it makes no sense to make the structure of the decoder layer a recurrent form. We should further note that in the case without the participation of storage installations, outputs of each time steps are determined only on their own states. Therefore, different from the typical structure of a transformer decoder, we only conduct an attention mechanism between output and input without the masked self-attention among outputs. 

To calculate the correlation between nodes and links, we defined the latent state of links in the graph using their positional information via the positional encoding of the two nodes connected by the link. The positional encoding of the link between node $i$ and $j$ was denoted as follows:
\begin{equation}
    \mathbf{P}_{\textrm{link}}=\left\{ \mathbf{p}_{\textrm{link},l}|l:\left ( i,j \right )\in \mathcal{L}\right\}\in \mathbb{R}^{2m\times L}
\end{equation}
\vspace{-15pt}
\begin{equation}
    \mathbf{p}_{\textrm{link},l} = \mathbf{p}_{\textrm{node},i}\|\mathbf{p}_{\textrm{node},j}
\end{equation}
where $\mathcal{L}$ denotes the set of all edges in the graph, which contains $L$ links. 

We treated NLAT layer as another function that maps the latent node states and link features to the new link expressions : 
$\mathfrak{G}_{\mathbf{W}_{\textrm{Q}}, \mathbf{W}_{\textrm{K}},\mathbf{W}_{\textrm{V}}}\,:\,\mathbb{R}^{KF'\times N}\,\to\,\mathbb{R}^{U\times L}$, where $\mathfrak{G}_{\mathbf{W}_{\textrm{Q}}, \mathbf{W}_{\textrm{K}},\mathbf{W}_{\textrm{V}}}$ denotes the NLAT layer parameterized by transformation matrices $\mathbf{W}_{\textrm{Q}}\in\mathbb{R}^{L\times 2m\times V}, \mathbf{W}_{\textrm{K}}\in\mathbb{R}^{N\times KF'\times V}$ and $\mathbf{W}_{\textrm{V}}\in\mathbb{R}^{N\times KF'\times U}$ for each links and nodes, $V$ is the length of last dimension of quires and keys matrices, $U$ denotes the number of latent features for each link. Note that each kink of transformation matrices is different for each link and nodes, in order to augment express power. We therefrom got a new expression for links $\mathbf{R}=\{ \mathbf{r}_{l}|l\in\mathcal{L}\}\in \mathbb{R}^{U\times L}$, column vector $\mathbf{r}_{l}\in\mathbb{R}^{U}$ denotes output expression for link $l$.

We calculated the correlation between link $l$ and nodes $i$ by applying a scaled dot product:
\begin{equation}
    \hat{e}_{l,i}=\frac{\left ( \textbf{W}^{\intercal}_{\textrm{Q},l}\mathbf{p}_{\textrm{link},l} \right )^{\intercal}\cdot \textbf{W}^{\intercal}_{\textrm{K},i}\mathbf{h}''_{i}}{\sqrt{V}}
\end{equation}
\vspace{-15pt}
\begin{equation}
    \hat{\alpha }_{l,i}=\mathrm{softmax}_{i}\left ( \hat{e}_{l,i} \right ) = \frac{\mathrm{exp}(\hat{e}_{l,i})}{\sum_{k}\mathrm{exp}(\hat{e}_{l,k})}
\end{equation}
where $\hat{e}_{l,i}$ denotes the node-link attention coefficient of link $l$ while focusing on the node $i$, and 
$\textbf{W}_{\textrm{Q},l} \in\mathbb{R}^{2m\times V}$,  
$\textbf{W}_{\textrm{K}, i} \in\mathbb{R}^{KF'\times V}$ denote the linear transformation query matrix and key matrix for link $l$ and node $i$ respectively. Then we conducted aggregation by weighted summation yielding outputs:

\begin{equation}
    \mathbf{r}_{l}=\underset{i}{\sum}\hat{\alpha}_{l,i}\mathbf{W}_{\textrm{V},i}^{\intercal}\mathbf{h}''_{i}
\end{equation}

The output expression of links $\mathbf{R}$ is then fed into an MLP according to the final task. 

\subsection{MLP Output layer}
We then used MLP as the output layer, generating the amount of power dispatched in each link. The units in hidden layers are fully connected in MLP, it has $R$ hidden layers with $d_{k}$ hidden units per layer, $k=1,2,...,R$. Units inside the network are connected by weights and biases. 

We treated the MLP as a function mapping the high-dimensional features space of links to the output feature: $\mathfrak{H}_{\mathbf{W}',\mathbf{b}}\,:\,\mathbb{R}^{U\times L}\,\to\,\mathbb{R}^{L}$, where $\mathfrak{H}_{\mathbf{W}',\mathbf{b}}$ denotes the MLP layer parameterized by weight matrix $\mathbf{W}'=\{ \mathbf{W}'_{k}\in\mathbb{R}^{d_{k-1}\times d_{k}}|k=1,2,...,R\}$ and bias $\mathbf{b}=\{\mathbf{b}_{k}\in\mathbb{R}^{d_{k}}|k=1,2,...,R \}$. Note that the input link expressions are $\mathbf{R}\in \mathbb{R}^{U\times L}$, we fed the expression for each nodes $\mathbf{r}_{l}\in\mathbb{R}^{U}$ into a same MLP network, thus $d_{0}=U$, and the outputs of MLP are power dispatches $\mathbf{F}\in\mathbb{R}^{L}$, thus the output layer has the weight matrix $\mathbf{W}'_{\textrm{out}}\in\mathbb{R}^{d_{K}\times 1}$ and bias $b_{\textrm{out}}\in\mathbb{R}$.

We denoted the output of each hidden layer inside the network as $\mathbf{h}_{k}\in\mathbb{R}^{d_{k}}$, which was also the input to the next hidden layer. Thus, for each hidden layer, we expressed the data flow as follows:
\begin{equation}
    \mathbf{h}_{k}=\textrm{LeakyReLU}(\mathbf{W}'_{k}\cdot \mathbf{h}_{k-1}+\mathbf{b}_{k})
\end{equation}
\vspace{-15pt}
\begin{equation}   \mathbf{h}_{0}=\mathbf{h}_{\textrm{in}}=\textbf{R}\quad\mathbf{h}_{\textrm{out}}=\textrm{Tanh}(\mathbf{W}'_{\textrm{out}}\cdot \mathbf{h}_{R}+b_{\textrm{out}})
\end{equation}
where $\textrm{LeakyReLU}$ and $\textrm{Tanh}$ are activation functions, $\textbf{R}$ corresponds to the output link expression from NLAT layer. $\mathbf{h}_{\textrm{out}}\in\mathbb{R}^{1\times L}$ correspond to the normalized power dispatches, which are the output of OPF problem.

\section{Experiments} \label{Sec4}

We started the experiments by generating training datasets, then applied the proposed machine learning framework in two scenarios. In the first scenario, we simplified the European power grid that each country was represented by a single node and countries were connected by links representing today’s topology. In the second scenario, we increased the number of total nodes in the power grid, which corresponds to a more complex scenario. Then we trained the proposed attention-based neural network on both cases, analyzed the performance and compared it to other methods.  We used PyPSA-Eur which is a sector-coupled open optimisation model of the European energy system, to build up the power system model and download system parameters and weather data. We then clustered the power grid which initially had thousands of nodes, leaving the number of nodes that we needed. At last, to make the training easier, we normalized the input data and labels as preprocessing. 

\subsection{Dataset generation}
A dataset with inputs and labels is required for data-driven supervised learning. The parameters in the power system model as well as the electricity consumption data and weather data are acquired through an open source tool, PyPSA-Eur \cite{PyPSAEur}, which is a powerful toolbox for simulating and optimizing modern power system. PyPSA-Eur provides a multifunctional power grid model including many kinds of energy sectors.

\subsubsection{Power system model}
\begin{figure}[h]
    \begin{center}
        \includegraphics[width=7.5cm]{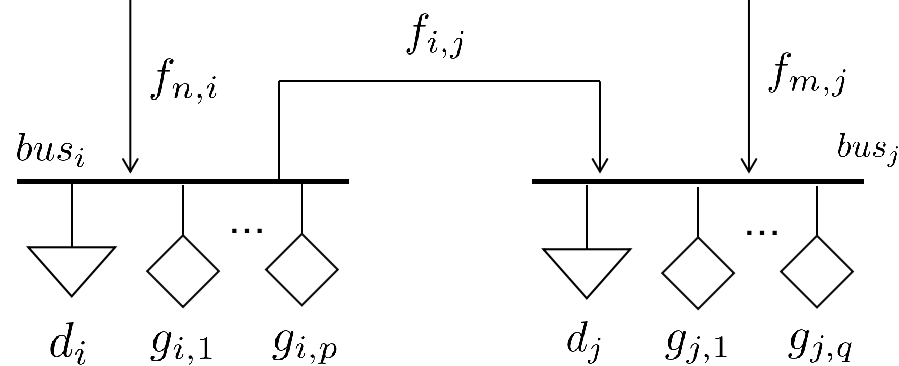}
    \end{center}
    \caption{A simplified case of power grid model inside PyPSA-Eur.}
    \label{fig:pypsa_model}
\end{figure}

Within PyPSA-Eur, the power grid is represented by nodes connected by edges. Each node signifies a bus, housing various types of generators and electricity consumption units. An edge linking two nodes represents a power transmission line, with a simplified illustration shown in Fig.~\ref{fig:pypsa_model}. The electricity consumption unit at bus $i$ and bus $j$ is denoted by $d$, while generators are denoted by $g$ with energy carrier $p/q$. The term ${f}_{i,j}$ denotes the power dispatch with the direction pointed from bus $i$ to bus $j$. 

In order to streamline the OPF problem, we omitted storage technologies in the scenarios discussed in this paper. Consequently,  the states of the power grid at each timestep are independent, relying solely on the input data at that specific timestep, i.e., the demand and weather data. Without losing generality, we chose Open Cycle Gas Turbine (OCGT) and hard coal power plants as conventional generators and chose onshore wind turbines and photovoltaic panels as renewable generators. Key parameters are detailed in Appendix \ref{appendix:para}.

\subsubsection{Clustering and optimization}
\begin{figure*}[h!]
    \begin{center}
        \begin{subfigure}{0.45\textwidth}
        \includegraphics[width=\linewidth]{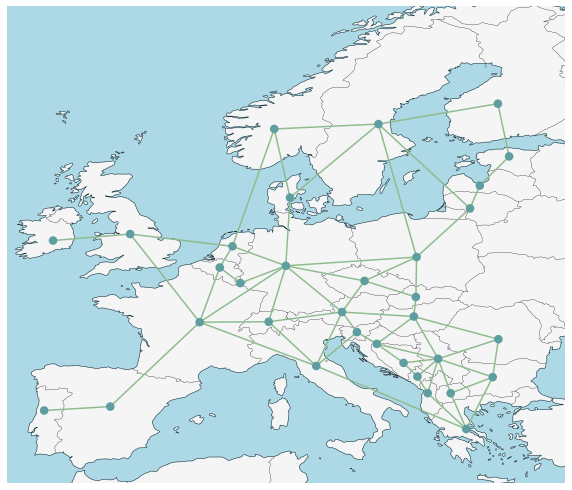}
        \caption{} 
        \label{fig:33_sys_wocap_a}
        \end{subfigure}%
        \hspace*{0cm}   
        \begin{subfigure}{0.45\textwidth}
        \includegraphics[width=\linewidth]{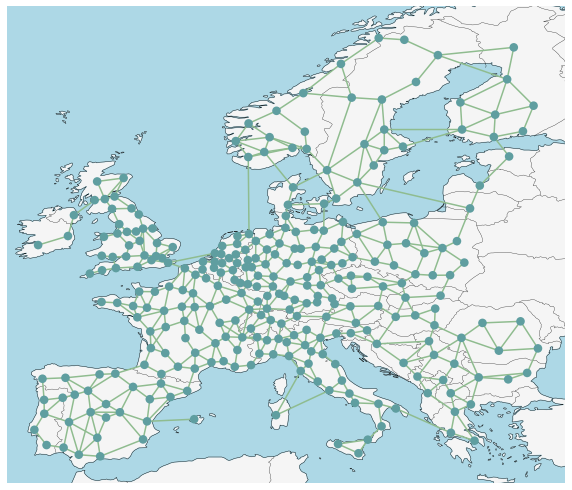}
        \caption{} 
        \label{fig:33_sys_wocap_b}
        \end{subfigure}
    \end{center}
    \caption{The layout of clustered power system.}
    \label{fig:two_sys_wocap}
\end{figure*}

The original power system model comprises 5400 nodes. We employed the clustering function in PyPSA-Eur to condense it using the k-nearest neighbors (KNN) algorithm, resulting in two distinct scenarios for the simplified power system. The first one has 33 nodes and 60 links, with each country corresponding to a node, as depicted in Fig.~\ref{fig:33_sys_wocap_a}. In the second scenario, the simplified power system has 300 nodes and 550 links, which is shown in Fig.~\ref{fig:33_sys_wocap_b}. Detailed descriptions of the power grid structure in both scenarios are available in Appendix~\ref{appendix:cluster}. It's worth noting that nodes are heterogeneous, each contains varying numbers of generators after clustering, with some even lacking local generators and solely consuming energy imported from other nodes.

After clustering, we optimized the power networks in each scenario via PyPSA-Eur. There are interfaces for different optimization solvers integrated into the PyPSA-Eur API, such as Gurobi, which addresses model-based optimization problems by using conventional methods, e.g., the interior point method. We subsequently invoked Gurobi to solve this linear optimal power flow problem to get the optimal results, which served as labels for training the neural networks.

\subsubsection{Data preprocessing}
%Either the input power consumption and weather data or the output optimal power dispatches and generator setting points all have different dynamic ranges. 
Both the input data, which includes power consumption and weather information, and the output data, encompassing optimal power dispatches and generator setting points, possess distinct dynamic ranges.
To ease the training process and improve the performance of the ML algorithm, we employed linear scaling to normalize these data as a data preprocessing step:
\begin{equation}   \tilde{P}_{j}^{\textrm{D}}=P_{j}^{\textrm{D}}/P_{j}^{\textrm{D}_{\textrm{max}}}
\end{equation}
\vspace{-15pt}
\begin{equation}
    \tilde{F}_{l}=F_{l}/\bar{F}_{l}
\end{equation}
with $j\in\mathcal{N}$ indexing the set of nodes, and $l\in\mathcal{L}$ labeling the set of links. $\tilde{P}_{j}^{\textrm{D}}\in \left [ 0, 1 \right ]$ denotes normalized active power demand at node $j$, and $\tilde{F}_{l}\in \left [ -1, 1 \right ]$ denotes normalized active power dispatch. $\bar{F}_{k}$ equals to the nominal power showed in \eqref{eq_2_act_flow_lim}. $P_{j}^{\textrm{D}_{\textrm{max}}}$ denotes the statistical maximum power demand according to all historical data. Note that another input feature, $\boldsymbol{\eta}$, representing the power capacity coefficient, inherently falls within the range of $\left [ 0, 1 \right ]$, eliminating the need for normalization.

\subsection{Machine Learning Strategy}
Neural networks can be trained in a supervised way, which can be seen as constructing a mapping between input data and output. Using backpropagation algorithm, the weights and biases between layers are adjusted based on the derivative of loss, continuing until a predetermined large number of training iterations, i.e., epochs, are met. After the training process, the neural network's performance can be evaluated with proper metrics, tailored to the nature of the problem or the form of network output.

\subsubsection{General mathematical expression}
Now we first present the general mathematical representation of the machine learning task. It can be expressed as an optimization problem aimed at searching the optimal model parameters, in the meanwhile giving the normalized power dispatches and total power generation at each node:
\begin{equation}
\begin{aligned}
    \underset{\hat{P}_{j}^{\textrm{G}_{\textrm{total}}},\: \hat{F}_{l},\: \boldsymbol{\lambda }}{\textrm{minimize}} \quad \frac{1}{L} &\sum_{l\in\mathcal{L}}\textrm{LogCosh} ( \hat{F}_{l}-\tilde{F}_{l} ) \quad+\\ \alpha \cdot \frac{1}{N}\sum_{j\in\mathcal{N}} &\textrm{MSE}( P_{j}^{\textrm{G}_{\textrm{total}}}-\tilde{P}_{j}^{\textrm{D}}\cdot P_{j}^{\textrm{D}_{\textrm{max}}}- \sum_{\substack{l:(j,k)\\k\in\mathcal{N}^{j}}} \hat{F}_{l}\cdot \bar{F}_{l} )
\end{aligned}
\label{eq_4B_lostfunc}
\end{equation}
subject to
\begin{equation}
    \hat{P}_{j}^{\textrm{G}_{\textrm{total}}} = \tilde{P}_{j}^{\textrm{D}}\cdot P_{j}^{\textrm{D}_{\textrm{max}}} + \sum_{\substack{l:(j,k)\\k\in\mathcal{N}^{j}}} \hat{F}_{l}\cdot \bar{F}_{l}
    \label{eq_4B_total_power_det}
\end{equation}
\vspace{-15pt}
\begin{equation}
    0\leq \hat{P}_{j}^{\textrm{G}_{\textrm{total}}}\leq \bar{P}_{j}^{\textrm{G}}
    \label{eq_4B_total_power_lim}
\end{equation}
\vspace{-15pt}
\begin{equation}
    -1\leq \hat{F}_{j}\leq 1
    \label{eq_4B_flow_lim}
\end{equation}
where $\hat{F}_{l}$ and $\hat{P}_{j}^{\textrm{G}_{\textrm{total}}}$ are predicted optimal power dispatch on link $l$ and total power generation at node $j$, respectively, calculated from \eqref{eq_3_projection}. 
$\tilde{F}_{l}$ is the normalized label for neural network output, i.e., the power dispatch. ${\boldsymbol\lambda}$ is the array of neural network parameters. $\bar{P}_{j}^{\textrm{G}}$ is the nominal power of node $j$ which is the sum of nominal power of all generators at that node. $\eta_{j,n} \in \left [ 0,\:1 \right ] $ denotes the capacity coefficient related to the weather condition, i.e., wind speed and solar radiation. $\textrm{LogCosh}$ denotes the logarithm of hyperbolic cosine function, $\textrm{MSE}$ corresponds to the mean absolute error function.

The loss function \eqref{eq_4B_lostfunc} has two terms, the first term is residual loss measuring the difference between predictions and labels, and the second term is a penalty measuring the violation of nodal balance, with a scaling factor $\alpha$ adapts the order of magnitude of the penalty term to be the same as the residual term. We added the second term to the loss function to force the neural network to learn the nodal balance. According to nodal balance constraint \eqref{eq_4B_total_power_det}, the total power generation at each node is determined by the power dispatch related to that node, therefore the only independent variable in this problem is $\hat{F}_{l}$, and the only two constraints that we need to consider are \eqref{eq_4B_total_power_lim} and \eqref{eq_4B_flow_lim}.

\subsubsection{Fulfillment of Constraints}
In most instances, training neural networks involves solving an unconstrained optimization problem. Without proper constraints, we might encounter scenarios where power dispatches or generators exceed their capacities. 

One way to incorporate the constraints is using Lagrangian relaxation or its variants \cite{fioretto2020predicting}, and also introducing penalty terms as we did for the nodal balance, both of which aim to embed constraints into the learning loss function, rendering the problem constraint-less. %Although these methods are able to encourage the satisfaction of constraints, the neural network output is not ensured to be feasible. 
While these methods promote constraint satisfaction, they don't guarantee a feasible output from the neural network.
An alternative is to modify the neural network directly, such as employing specific activation functions. In our case, we selected the $\textrm{Tanh}$ activation function for the last layer of the MLP:
\begin{equation}
    \textrm{Tanh}(x)=\frac{e^{x}- e^{-x}}{e^{x}+ e^{-x}}
\end{equation}
Since $\textrm{Tanh}(x) \in (-1,\: 1)$, it makes the constraint \eqref{eq_4B_flow_lim} strictly satisfied. However, it's still possible to violate other constraints, thus we introduced a post-processing giving a strictly feasible solution.

\subsubsection{Post-processing}
Infeasible operating parameters for the power grid are fatal since they potentially lead to grid failure and even outage. The post-processing is designed to identify a feasible solution, given the produced optimal power dispatch $\hat{\textbf{F}} = \{\hat{F}_{l}| l\in \mathcal{L}\}$ and the various constraints that must be satisfied.

Assuming the actual solution is proximate to $\hat{\textbf{F}}$, we projected $\hat{\textbf{F}}$ onto the surface of the polyhedral feasible space \cite{pan2020deepopf}, since DCOPF is a linear optimization problem, the optimal solution lies on the surface of this feasible space. We conducted the projection by solving a convex quadratic optimization problem searching for the optimal solution nearest to $\hat{\textbf{F}}$. The quadratic optimization problem is formulated as follows:
\begin{equation}
    \underset{\dot{\textbf{F}} }{\textrm{min}}\; \left\| \hat{\textbf{F}}-\dot{\textbf{F}} \right\|^2\quad \textrm{s.t.}\; \; \dot{\textbf{F}} \: \textrm{satisfies}\:\eqref{eq_4B_total_power_det}\sim\eqref{eq_4B_flow_lim}
    \label{eq_4B_projection}
\end{equation}
where $\dot{\textbf{F}}= \{\dot{F}_{l}| l\in \mathcal{L}\}\in \mathbb{R}^{L}$ denotes the solution after projection. The problem in \eqref{eq_4B_projection} is straightforward and can be quickly solved. The optimal power dispatch $\dot{\textbf{F}}$ is thus ensured to be feasible and ready for subsequent use.

\subsubsection{Evaluation Metric}
Various metrics can be used to measure the goodness of prediction for regression tasks, e.g., coefficient of determination which is also known as $\textrm{R}^2$, it provides a measure of how well the observed outcomes are replicated by the model, based on the proportion of total variation of outcomes explained by the model \cite{carpenter1960principles}. However, it can be confusing when the variance of predicted data is very small, that $\textrm{R}^2$ tends to $-\infty $. This can happen in OPF problem when there is always congestion on some links. Therefore, we use in this work the mean arctangent absolute percentage error (MAAPE) \cite{kim2016new}, defined as follows:
\begin{equation}
    \textrm{MAAPE}=\frac{1}{N}\sum_{i=1}^{N}\textrm{arctan}\left ( \left| \frac{A_i-F_i}{A_i}\right| \right )
\end{equation}
where there are $N$ data points, $A_i$ and $F_i$ denote $i$-th actual and forecast values, $\textrm{arctan}$ is the arctangent function, and it's output range is $\left [ 0,\pi/2 \right )$. The arctangent function prevents the metrics from reaching infinity. The less the MAAPE, the better the regression.

\subsection{Generator Output Calculation}
\begin{algorithm}[bp!]
\SetAlgoLined
\caption{Merit order}\label{alg:MOE}
    \KwData{$\hat{P}_{j}^{\textrm{G}_{\textrm{total}}},\; c_{i,j},\; \bar{P}_{j,n}^{\textrm{G}},\; RANK$}
    \KwResult{$\hat{P}_{j,n}^{\textrm{G}}$}
    $\delta \gets 1e^{-3}$\;
    \For{$j\in \mathcal{N}$}{
        \If{$\mathcal{G}_{j}\notin \varnothing$}{
            $P_{j}^{\textrm{G}_{\textrm{left}}} \gets \hat{P}_{j}^{\textrm{G}_{\textrm{total}}}$ \;
            \For{$n \in RANK$}{
                \If{$P_{j}^{\textrm{G}_{\textrm{left}}} > \delta$}{
                    \If{$n \in \mathcal{G}_{j}$}{
                        \eIf{$P_{j}^{\textrm{G}_{\textrm{left}}}> c_{i,j}\cdot \bar{P}_{j,n}^{\textrm{G}}$}{
                            $\hat{P}_{j,n}^{\textrm{G}} \gets c_{i,j}\cdot \bar{P}_{j,n}^{\textrm{G}}$\;
                            $P_{j}^{\textrm{G}_{\textrm{left}}} \gets P_{j}^{\textrm{G}_{\textrm{left}}} - c_{i,j}\cdot \bar{P}_{j,n}^{\textrm{G}}$
                        }{
                            $\hat{P}_{j,n}^{\textrm{G}} \gets P_{j}^{\textrm{G}_{\textrm{left}}}$\;
                            $P_{j}^{\textrm{G}_{\textrm{left}}} \gets 0$
                        }
                    }
                }
            }
    }
}
\end{algorithm}
Given the optimal power dispatches $\dot{\textbf{F}}$ and the subsequent calculated total power output of each node $\hat{\mathbf{P}}^{\textrm{G}_{\textrm{total}}}=\{\hat{P}_{j}^{\textrm{G}_{\textrm{total}}}|j\in\mathcal{N}\}$, we then calculated the setting point i.e., the output power for each generators according to merit order. The merit order is a way to rank the source of energy given their marginal price, the cheapest energy first and the most expensive energy last. Given the actual marginal cost of each kind of generator listed in Appendix \ref{appendix:para}, we thus ranked each energy type as: solar, wind, OCGT and coal. The generator outputs are limited as follows:
\begin{equation}
    0\leq \hat{P}_{j,n}^{\textrm{G}}\leq c_{j,n}\cdot\bar{P}_{j,n}^{\textrm{G}} 
    \label{eq_4C_gen_lim}
\end{equation}
\vspace{-15pt}
\begin{equation}
c_{j,n}=
\begin{cases}
     \eta_{j,n}&\; n\in \mathcal{G}_{j}^{\textrm{r}}\\
     1 &\;  n\in \mathcal{G}_{j}^{\textrm{c}}
     \label{eq_4C_weather_coe}
\end{cases}
\end{equation}
where $\hat{P}_{j,n}^{\textrm{G}}$ and $\bar{P}_{j,n}^{\textrm{G}}$ denote the predicted optimal active power and nominal active power of the generator $n$ at node $j$, $c_{j,n}$ denotes the capacity coefficient. $n\in\mathcal{G}_{j}$ which is the set of all generators at node $j$, $\mathcal{G}_{j}=\mathcal{G}_{j}^{\textrm{r}}\cup \mathcal{G}_{j}^{\textrm{c}}$, where $\mathcal{G}_{j}^{\textrm{r}}$ and $\mathcal{G}_{j}^{\textrm{c}}$ denote the set of renewable generators and conventional generators at node $j$, respectively. Constraints \eqref{eq_4C_gen_lim} and \eqref{eq_4C_weather_coe} means that the active power output of conventional generators is limited between $0$ and their nominal power, however the maximum output of renewable generators is influenced by a coefficient $\eta$ related to weather conditions. We outline the process of power output calculation in the following Algorithm \ref{alg:MOE}.

Thus, the active power output of each generator at each node is determined sequentially according to each one's marginal price from lowest to highest until the total amount is reached. In such way determining the power output of generators indirectly, the nodal balance constraint \eqref{eq_4B_total_power_det} is naturally satisfied, and since weather condition and nominal power of each generator are considered as known input, constraint \eqref{eq_4C_gen_lim} is also satisfied during the power allocation.

\begin{figure*}[t!]
    \centering
    \rotatebox[origin=c]{90}{First component} 
    \hspace{0.1cm}
    \begin{subfigure}[c]{.32\textwidth} 
        \centering
        Attention window 1
        \par\smallskip
        \includegraphics[width=\linewidth]{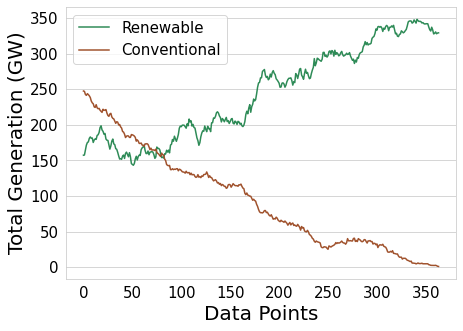}
        \vspace{-.6cm}
        \caption{}
        \label{fig:PCA_trend_w1_1}
    \end{subfigure}
    \hspace{-0.15cm}
    \vspace{.5cm}
    \begin{subfigure}[c]{.32\textwidth}
        \centering
        Attention window 2
        \par\smallskip
        \includegraphics[width=\linewidth]{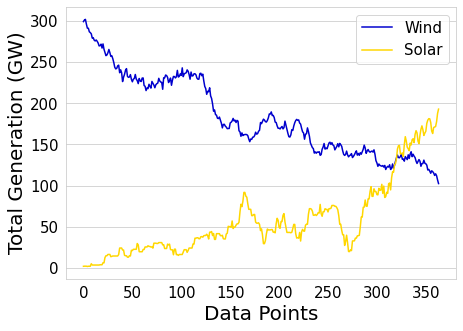}  
        \vspace{-.6cm}
        \caption{}
        \label{fig:PCA_trend_w2_1}
    \end{subfigure} 
    \hspace{-0.15cm}
    \begin{subfigure}[c]{.32\textwidth}
        \centering
        Attention window 3
        \par\smallskip
        \includegraphics[width=\linewidth]{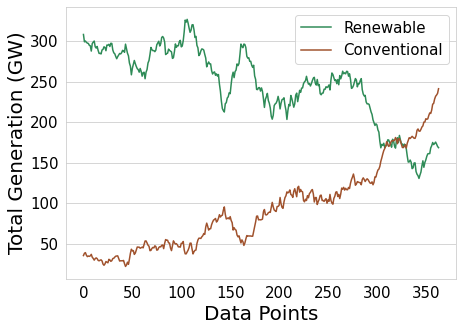} 
        \vspace{-.6cm}
        \caption{}
        \label{fig:PCA_trend_w3_1}
    \end{subfigure}
    \vskip -10pt
    \rotatebox[origin=c]{90}{Second component}
    \hspace{0.1cm}
    \begin{subfigure}[c]{.32\textwidth}
        \centering
        \includegraphics[width=\linewidth]{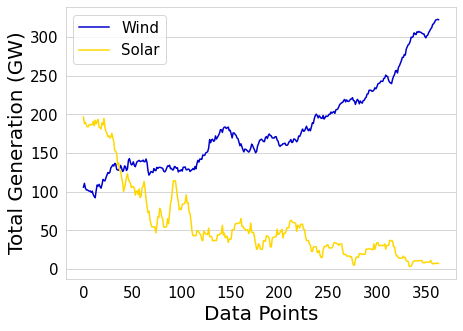}  
        \vspace{-.6cm}
        \caption{}
        \label{fig:PCA_trend_w1_2}
    \end{subfigure}
    \hspace{-0.15cm}
    \begin{subfigure}[c]{.32\textwidth}
        \centering
        \includegraphics[width=\linewidth]{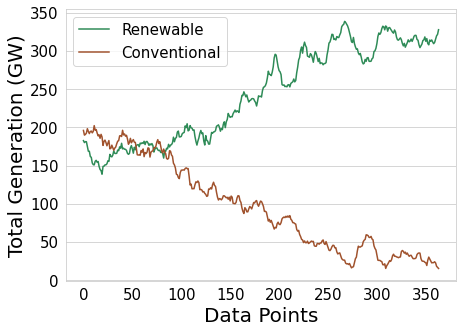} 
        \vspace{-.6cm}
        \caption{}
        \label{fig:PCA_trend_w2_2}
    \end{subfigure}
    \hspace{-0.15cm}
    \begin{subfigure}[c]{.32\textwidth}
        \centering
        \includegraphics[width=\linewidth]{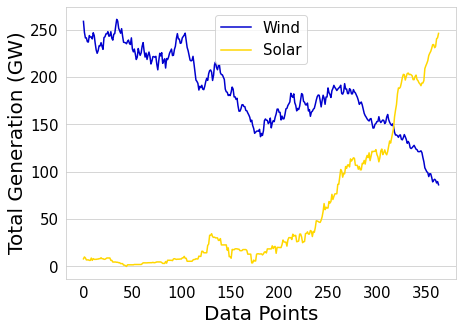}    
        \vspace{-.6cm}
        \caption{}
        \label{fig:PCA_trend_w3_2}
    \end{subfigure}
    \caption{Variation of power generation from different type of energy.}
    \label{fig:PCA_trend_node_att}
\end{figure*}

\section{Results and Discussions} \label{Sec5}

In this section, we analyze the experimental results from multiple perspectives. We considered two scenarios corresponding to the clustering results with 33 nodes left and 300 nodes left, respectively. We first test the effectiveness of our proposed attention-based neural networks by examining the attention matrix during two representative timesteps, each correlating to a distinct case. %PCA is applied then, and principal components are investigated to help explain the phenomenon appeared in each window inside the multi-window graph self-attention (SWM-GSAT) layer. 
Subsequently, the Principal Component Analysis (PCA) is employed to discern and study the primary components involved, shedding light on the observed phenomena within each window of the multi-window graph self-attention (SWM-GSAT) layer.
We further showcase predictions of the OPF results, gauging their accuracy through statistical evaluations. Lastly we compare the proposed attention-based neural network to other data-driven techniques, with the hyperparameters for all the methods listed in Appendix~\ref{appendix:hyper_para}.

\subsection{Effectiveness of Graph Self Attention}
For clarity and ease of presentation, we took the first scenario with 33 nodes as the example to elucidate the effectiveness of our proposed attention-based neural networks, the interpretability of larger scale power grid can be analogized to. First we applied PCA on the attention matrices given the whole test set, the first two components were picked to illuminate the underlying mechanism. Subsequently, we pinpointed two exemplary test cases for a detailed analysis of the attention matrix: Case 1, appeared during nighttime with strong wind, and Case 2, marked by daytime conditions with milder wind. 
Our selection of these two cases is based on the first component of the optimal power flow from PCA, which in overall aligns along the North-South direction \cite{hofmann2018principal}. Specifically, Case 1 exhibits a north-o-south power flow trend reflecting the absence of solar sources, while Case 2 demonstrates a south-to-north flow pattern indicating the limited wind sources.
%The reason why we chose these two cases is that the first component of optimal power flow in PCA has an overall orientation along the North-South \cite{hofmann2018principal}, thus we chose Case 1 who showed a trend of power flow from north to south, because of the lacking of solar source, and Case 2 who showed a trend of power flow from south to north, because of the lacking of wind source. 

In the SMW-GSAT layer, the attention matrix highlights the correlations among nodes in the graph, and the multi-window mechanism forces nodes in different windows to focus on different ranges of neighborhoods, as transmission efficiency and costs inherently limit the range of a node's influence. 

\begin{table}[h]
    \centering
    \renewcommand\arraystretch{1.3}
    \caption{Percentage variance of PCs}
    \begin{tabular}{@{}cccc@{}}
        \toprule
        Components & \makecell[c]{Window 1\\(\%)} & \makecell[c]{Window 2\\(\%)} & \makecell[c]{Window 3\\(\%)} \\ \midrule
        1st & 40.1 & 25.5 & 28.9 \\
        2nd & 19.9 & 19.2 & 16.2 \\
        3rd & 10.5 & 16.9 & 14.0 \\ \bottomrule
    \end{tabular}
    \label{tab:PC_variance}
\end{table}

\begin{figure*}[ht!]
    \centering
    \rotatebox[origin=c]{90}{First component}
    \hspace{0.1cm}
    \begin{subfigure}[c]{.32\textwidth}
        \centering
        Attention window 1
        \par\smallskip
        \includegraphics[width=\linewidth]{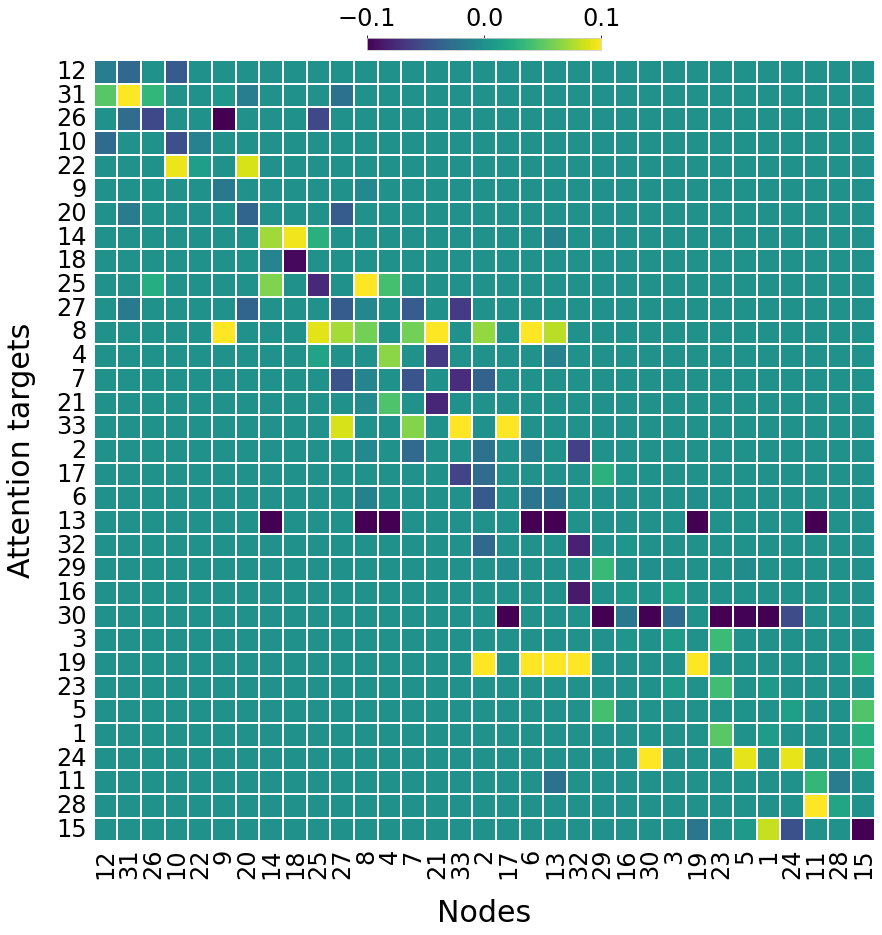}
        \vspace{-.6cm}
        \caption{}
        \label{fig:PCA_w1_1}
    \end{subfigure}
    \hspace{-0.15cm}
    \vspace{.5cm}
    \begin{subfigure}[c]{.32\textwidth}
        \centering
        Attention window 2
        \par\smallskip
        \includegraphics[width=\linewidth]{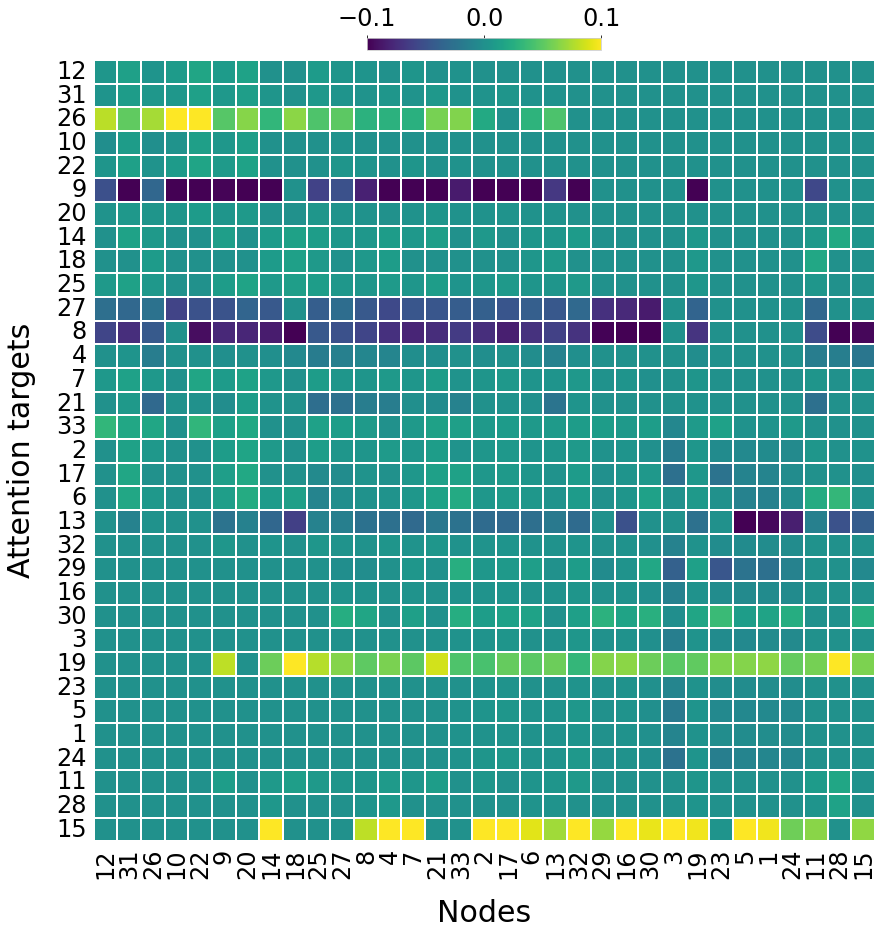}    
        \vspace{-.6cm}
        \caption{}
        \label{fig:PCA_w2_1}
    \end{subfigure} 
    \hspace{-0.15cm}
    \begin{subfigure}[c]{.32\textwidth}
        \centering
        Attention window 3
        \par\smallskip
        \includegraphics[width=\linewidth]{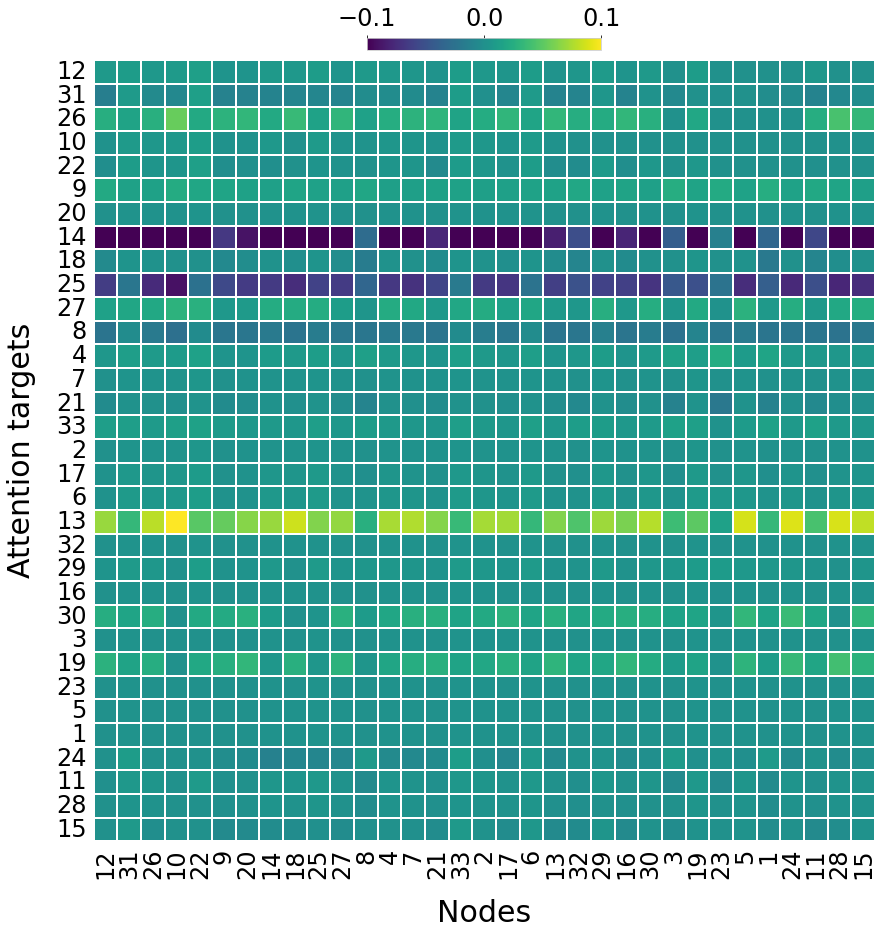}    
        \vspace{-.6cm}
        \caption{}
        \label{fig:PCA_w3_1}
    \end{subfigure}
    \vskip -10pt
    \rotatebox[origin=c]{90}{Second component}
    \hspace{0.1cm}
    \begin{subfigure}[c]{.32\textwidth}
        \centering
        \includegraphics[width=\linewidth]{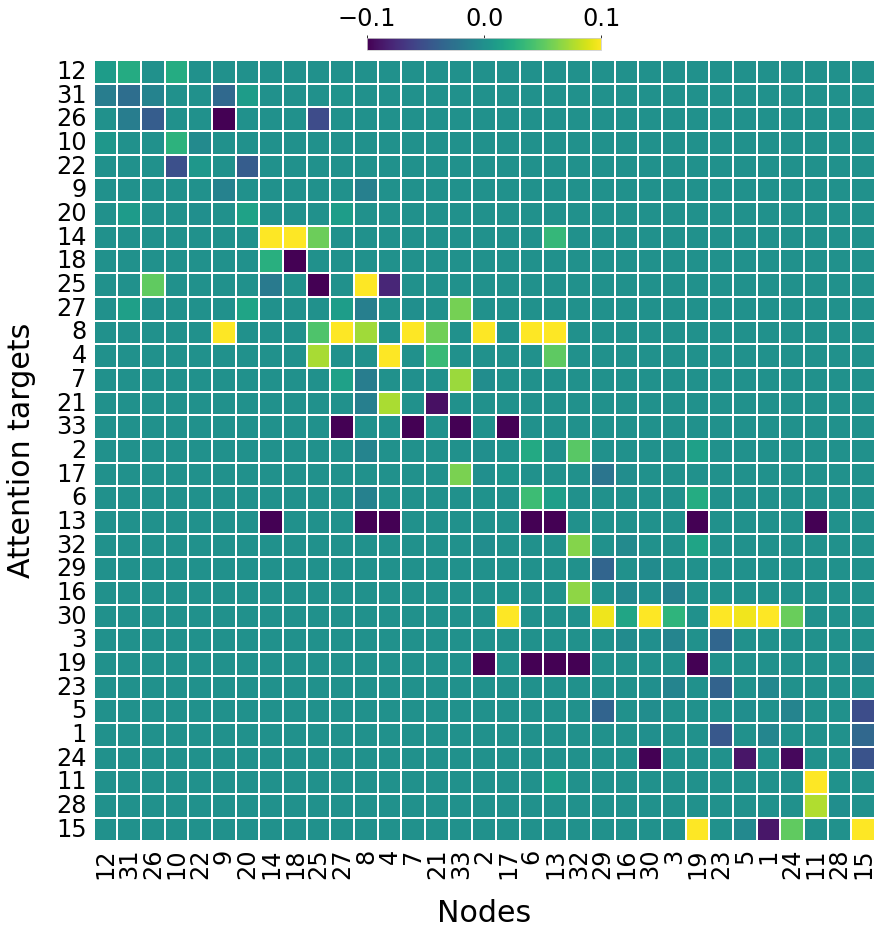}  
        \vspace{-.6cm}
        \caption{}
        \label{fig:PCA_w1_2}
    \end{subfigure}
    \hspace{-0.15cm}
    \begin{subfigure}[c]{.32\textwidth}
        \centering
        \includegraphics[width=\linewidth]{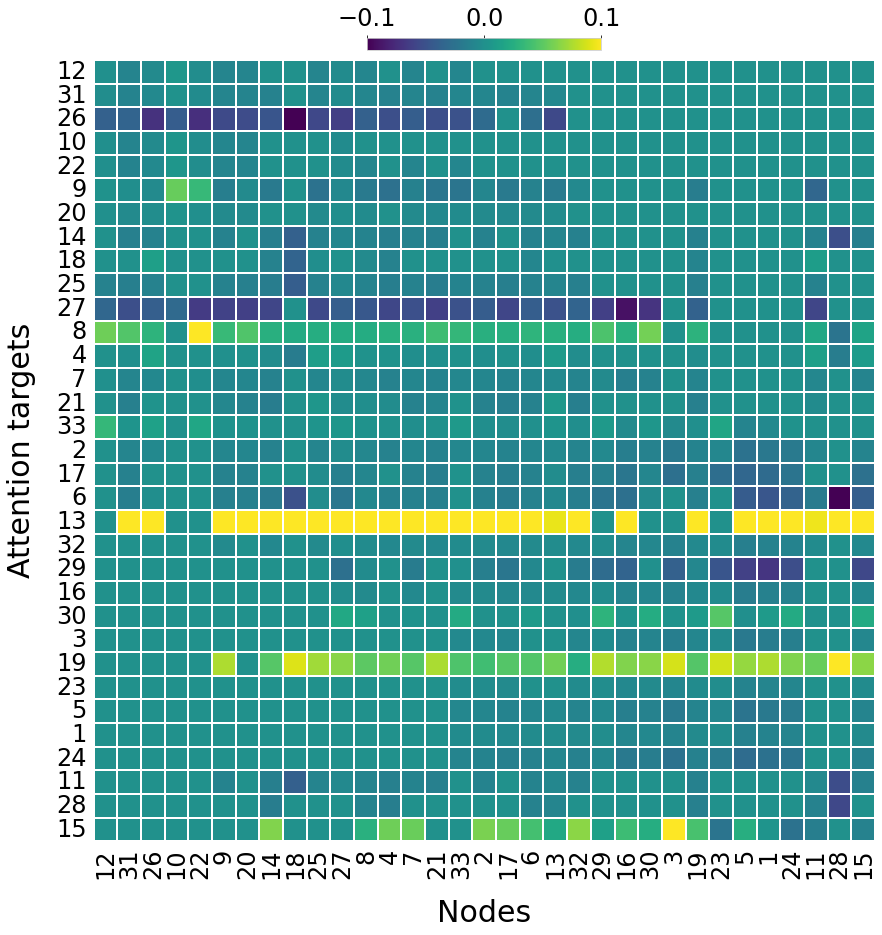}    
        \vspace{-.6cm}
        \caption{}
        \label{fig:PCA_w2_2}
    \end{subfigure}
    \hspace{-0.15cm}
    \begin{subfigure}[c]{.32\textwidth}
        \centering
        \includegraphics[width=\linewidth]{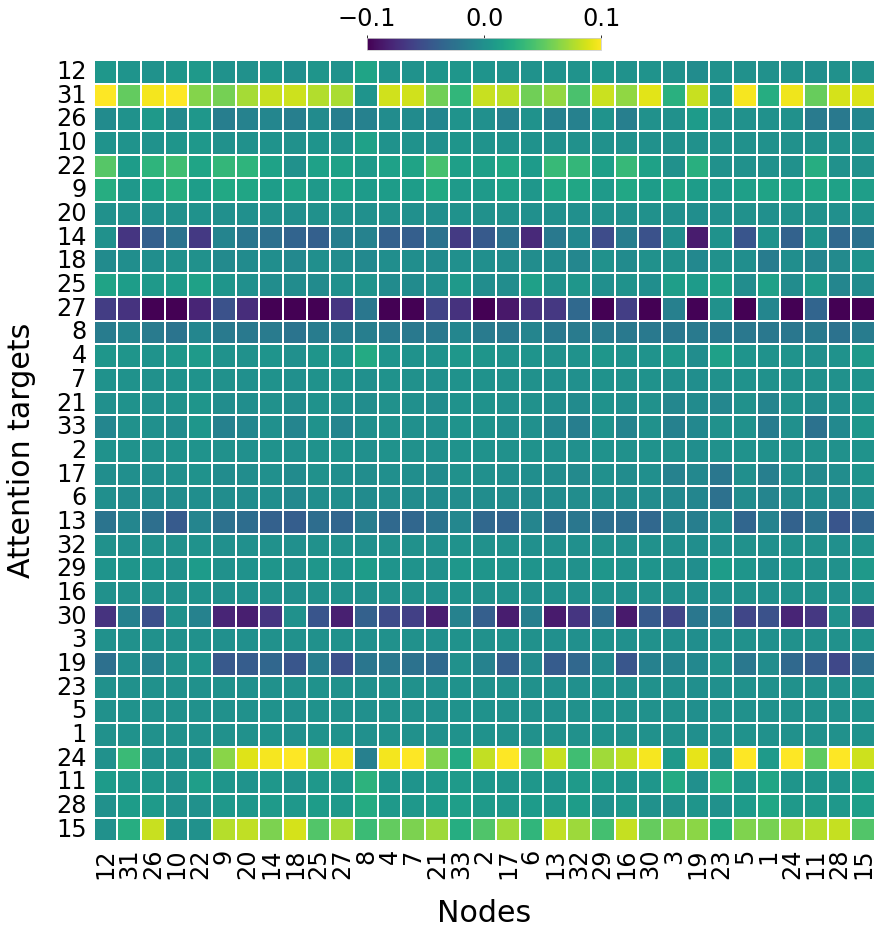}   
        \vspace{-.6cm}
        \caption{}
        \label{fig:PCA_w3_2}
    \end{subfigure}
    \caption{Principal components of node attention matrix. }
    \label{fig:PCA_node_att}
\end{figure*}

To better understand the meaning of the important nodes highlighted in the PCA results of the node attention matrix, we present variations of power generation from different types of energy sources in Fig.~\ref{fig:PCA_trend_node_att} that also reflect in which direction these principal components (PCs) distinguish the data. In Fig.~\ref{fig:PCA_trend_node_att} there were 4 kinds of statistical variables labeled in different colors, representing the total power generation from wind turbines, solar PV, and the aggregate power generation from renewable or conventional generations. In each sub-figure, the horizontal axis corresponds to the data points in the test set, and the vertical axis corresponds to the total power generation in GW. In different rows of Fig.~\ref{fig:PCA_trend_node_att}, data points are sorted by the transformed values in the direction of the first or second components of PCA from smallest to largest, and different columns correspond to the different PCA results from each three windows in the multi-window mechanism. Note that the lines in Fig.~\ref{fig:PCA_trend_node_att} were smoothed by an average filter. The PCs are depicted in Fig~\ref{fig:PCA_node_att}, in each figure, the horizontal axis identifies the node that is going to calculate the attention scores with other nodes, and the vertical axis designates the target nodes. The color intensity of each pixel corresponds to the attention score value, in other words, each column displays the attention scores between a node on the horizontal axis and other nodes on the vertical axis. The variance of these PCs is detailed in Table~\ref{tab:PC_variance}. To get a clearer observation of correlations, nodes are arranged according to their latitude from high to low, i.e., from north (No.12 Finland) to south (No.15 Greece). The node number and country comparison table can be checked in Appendix~\ref{appendix:cluster}. Following this, we present the analysis by combining the variation of power generation and PCs.

According to Fig.~\ref{fig:PCA_trend_w1_1}, the first principal component of the attention matrix in window 1 varies with respect to the variation of the proportion of renewable energy generation in the power grid. See Fig.~\ref{fig:PCA_w1_1}, since the smallest attention range determined by the mask only covers a few neighborhoods around each node, the attention matrix in window 1 highlights those nodes that are significant in their local area. Nodes that export large amounts of renewable energy for consumption by their neighborhoods get larger attentions, corresponding to the brighter points, such as node 8 (Germany) and node 19 (Italy). In contrast, when renewable resources are deficient, nodes such as 13 (France) that can export conventional energy or anticipate energy imports get large attention, as seen with the darker points. According to Fig.~\ref{fig:PCA_trend_w1_2}, the second component of the attention matrix in window 1 varies with respect to the alternation of wind energy and solar energy being the main energy sources. See Fig.~\ref{fig:PCA_w1_2}, nodes having significant solar energy capacity or those can export solar energy, like node 33 (Slovak Republic), node 13 (France), and node 19 (Italy), get larger attention when solar is the main energy source, as indicated by the darker points. Conversely, when wind energy is dominant, nodes with large wind energy capacities or those being able to export wind energy, such as node 14 (U.K.) and node 8 (Germany), are more emphasized in the attention matrix, as represented by the lighter points. Additionally, nodes highly dependent on solar energy, like node 30 (Serbia)--which only possesses solar generators--also get more attention.

According to Fig.~\ref{fig:PCA_trend_w2_1}, the first principal component of the attention matrix in window 2 varies with respect to the alternation of main energy sources from solar to wind. See Fig.~\ref{fig:PCA_w2_1}, the attention range spreads further by including more countries into the calculation of attention. Consequently, the first component in window 2 highlights nodes exerting influence both locally and across a broader spread. When solar energy predominates, important nodes include node 19 (Italy), node 15 (Greece), and node 26 (Norway) which have a more localized influence. On the contrary, when wind energy takes precedence, %important nodes include node 9 (Denmark) and node 27 (Poland), whoever node 8 (Germany) always meet the circumstance that needs to import power since low wind power generation which can't meet the power demand at itself, thus gets more attention. 
nodes 9 (Denmark) and 27 (Poland) stand out. Meanwhile, node 8 (Germany) consistently faces circumstances where its wind power generation falls short of its own power demands, leading it to import power and, consequently, getting more attention.
According to Fig.~\ref{fig:PCA_trend_w2_2}, the second component in window 2 varies with respect to the variation of the proportion of renewable energy generation in the power grid. See Fig.~\ref{fig:PCA_w2_2}, nodes exporting large amounts of renewable energy, such as node 13 (France) and node 19 (Italy),  get larger attentions, as represented by the brighter points. On the contrary, during times of scarcity in renewable resources, nodes like 27 (Poland) and 16 (Norway) which rely most on renewable energy get larger attention, depicted by the darker points.

According to Fig.~\ref{fig:PCA_trend_w3_1}, the first principal component of the attention matrix in window 3 varies with respect to the variation of the proportion of renewable energy generation. See Fig.~\ref{fig:PCA_w3_1}, the attention score was calculated among almost all the nodes in the power grid, thus the attention matrix in this window highlights nodes that influence the entire power grid the most. Two main renewable energy exporting nodes are highlighted, node 14 (U.K.) and node 25 (Netherlands), as seen by the darker points. On the contrary, when renewable resources are scarce, node 13 (France) takes precedence due to its substantial coal generation capacity, which tends to be more expensive than QCGT generators, making energy imports occasionally more cost-effective than activating coal power generators. %since it possesses the most capacity of coal generator which is more expansive than the OCGT generator, thus importing sometimes is cheaper than activating coal power generator. 
According to Fig.~\ref{fig:PCA_trend_w3_2}, the second component of the attention matrix in window 3 varies with respect to the alternation of main energy sources from wind to solar. See Fig.~\ref{fig:PCA_w3_2}, when solar is the main energy source, nodes exporting solar energy, such as node 31 (Sweden), node 24 (North Macedonia) and 15 (Greece), get more attention, marked by the lighter points. Conversely, when wind energy dominates, node 27 (Poland) gets larger attention since it's exporting wind energy. Simultaneously, node 14 (U.K.) gets attention since it's consuming energy, the same as node 30 (Serbia), as indicated by the darker points.

\begin{figure*}[h!]
    \centering
    \begin{subfigure}{.48\textwidth}
        \centering
        \includegraphics[width=\linewidth]{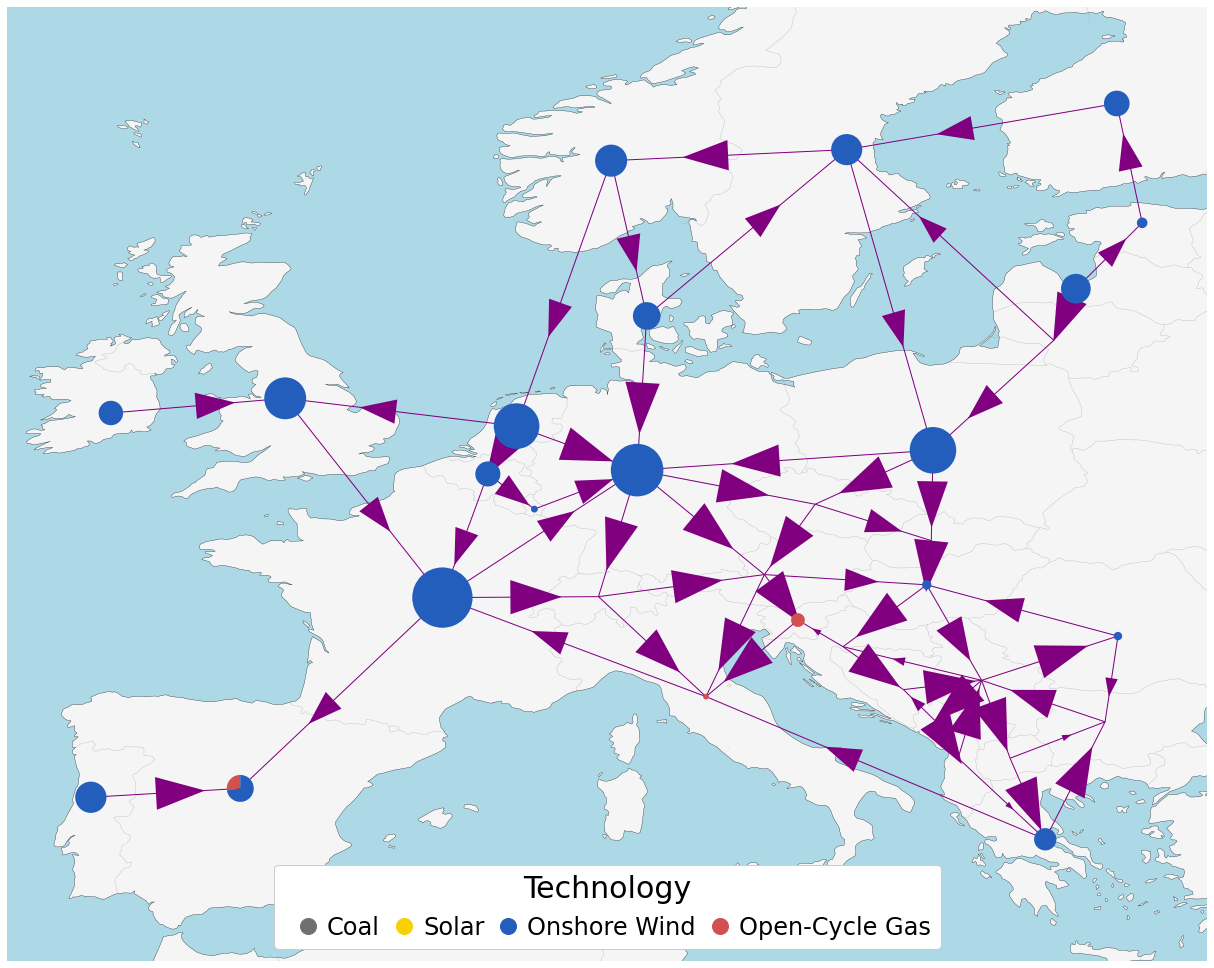}
        \caption{Case 1}
        \label{fig:optimal_powersys_108}
    \end{subfigure}
    \hspace{-0.1cm}
    \begin{subfigure}{.48\textwidth}
        \centering
        \includegraphics[width=\linewidth]{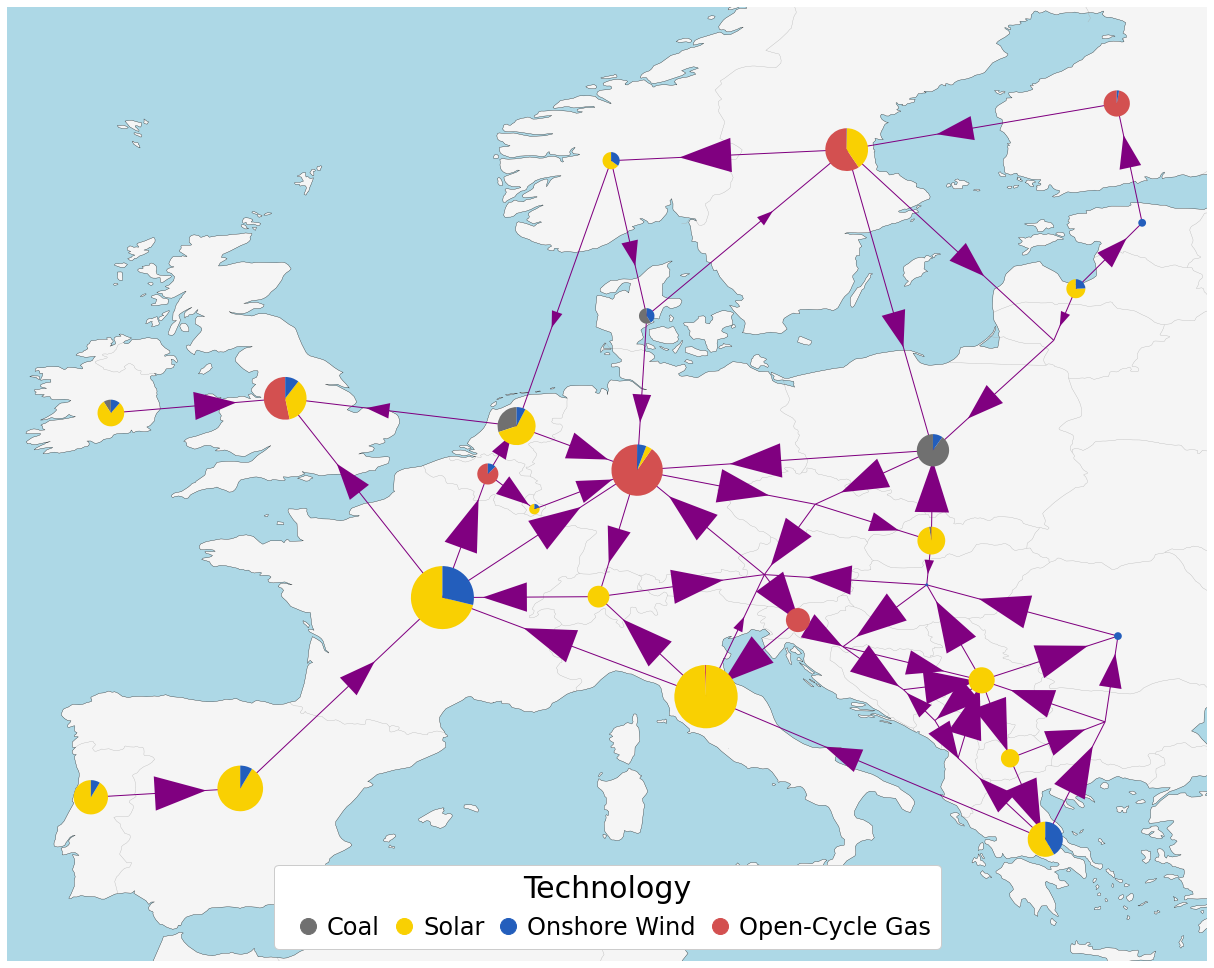}    
        \caption{Case 2}
        \label{fig:optimal_powersys_207}
    \end{subfigure} 
    \caption{Predicted optimal solutions for OPF.}
    \label{fig:optimal_solution_sys}
\end{figure*}

\begin{figure*}[h!]
    \centering
    \begin{subfigure}{.32\textwidth}
        \centering
        \includegraphics[width=\linewidth]{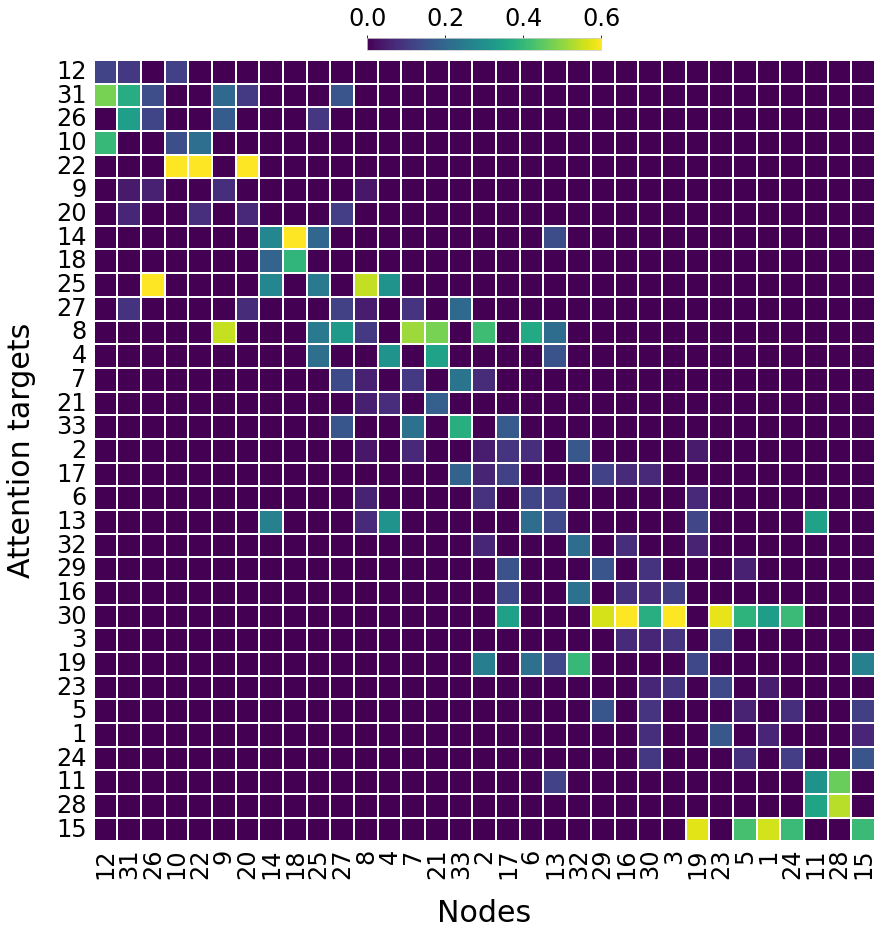}
        \caption{Attention window 1}
        \label{fig:node_att1_108}
    \end{subfigure}
    \hspace{-0.1cm}
    \begin{subfigure}{.32\textwidth}
        \centering
        \includegraphics[width=\linewidth]{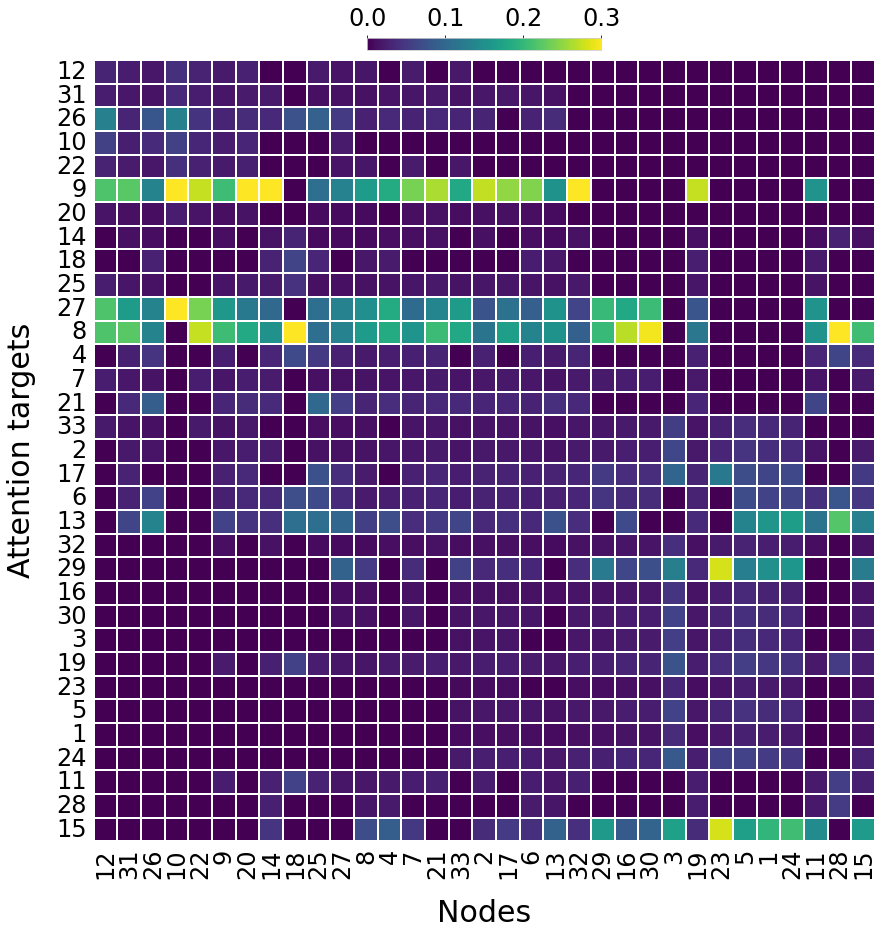} 
        \caption{Attention window 2}
        \label{fig:node_att2_108}
    \end{subfigure} 
    \hspace{-0.1cm}
    \begin{subfigure}{.32\textwidth}
        \centering
        \includegraphics[width=\linewidth]{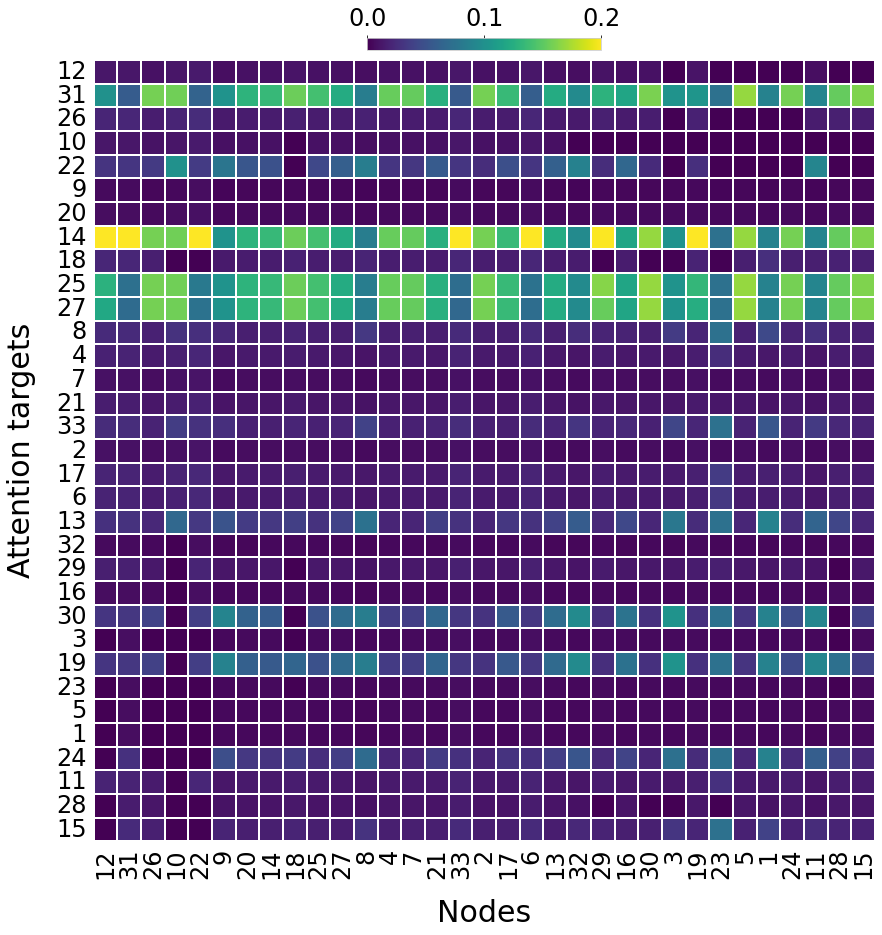} 
        \caption{Attention window 3}
        \label{fig:node_att3_108}
    \end{subfigure}
    \caption{Node attention matrix in Case 1.}
    \label{fig:node_attention_matrix_108}
\end{figure*}

\begin{figure*}[h!]
    \centering
    \begin{subfigure}{.32\textwidth}
        \centering
        \includegraphics[width=\linewidth]{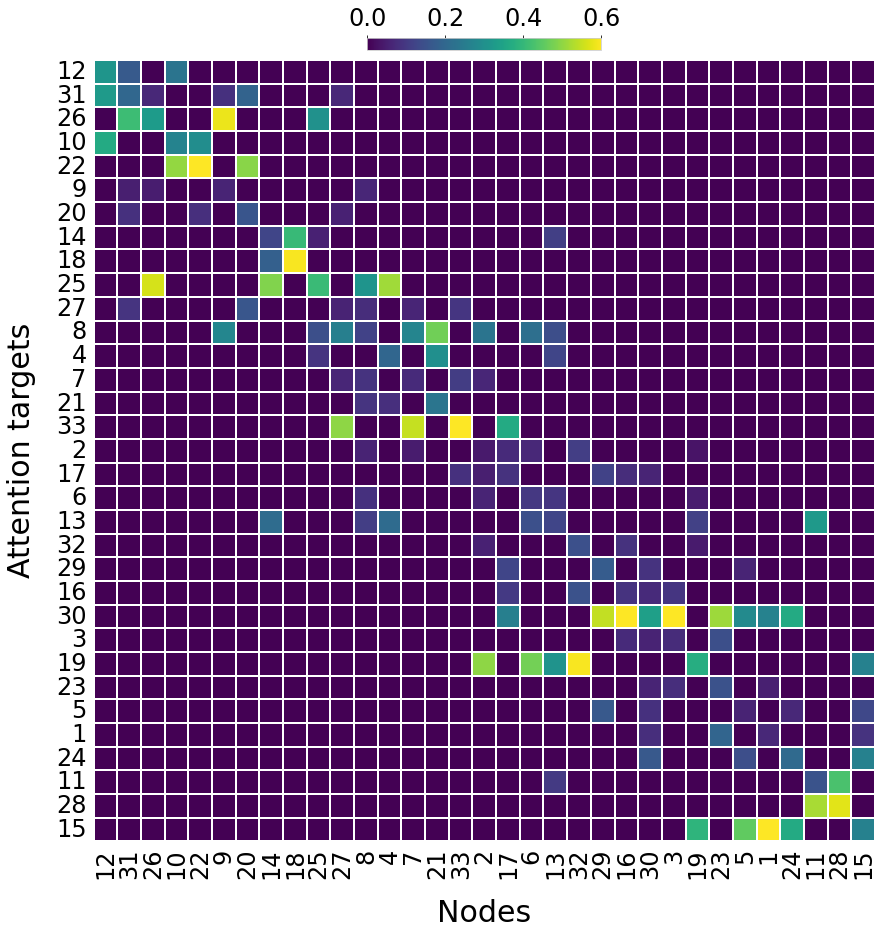}
        \caption{Attention window 1}
        \label{fig:node_att1_207}
    \end{subfigure}
    \hspace{-0.1cm}
    \begin{subfigure}{.32\textwidth}
        \centering
        \includegraphics[width=\linewidth]{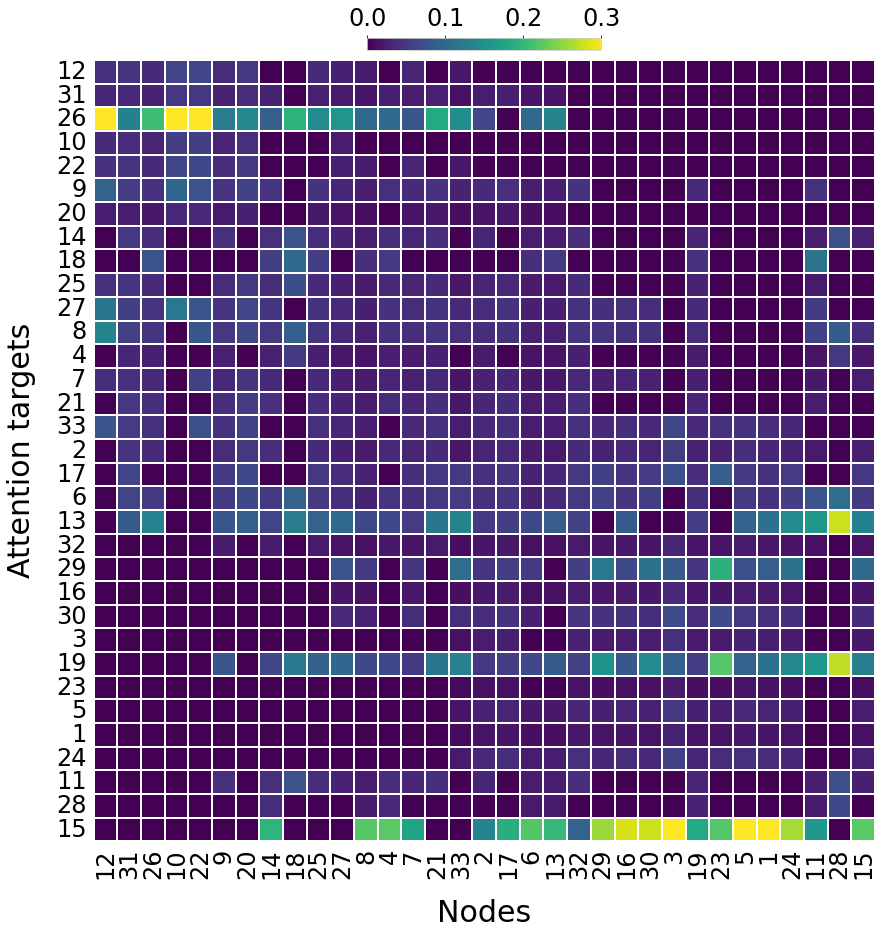}    
        \caption{Attention window 2}
        \label{fig:node_att2_207}
    \end{subfigure} 
    \hspace{-0.1cm}
    \begin{subfigure}{.32\textwidth}
        \centering
        \includegraphics[width=\linewidth]{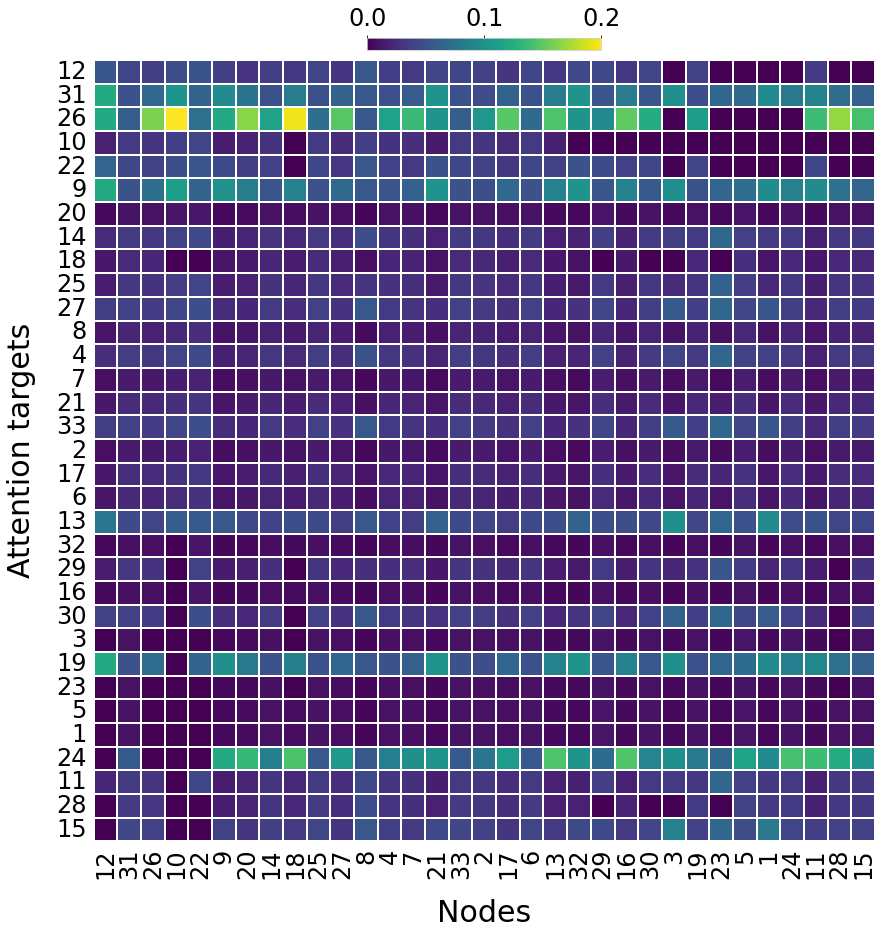}    
        \caption{Attention window 3}
        \label{fig:node_att3_207}
    \end{subfigure}
    \caption{Node attention matrix in Case 2.}
    \label{fig:node_attention_matrix_207}
\end{figure*}

To illustrate, we consider two representative cases. The optimal solutions for power dispatch and generation for these cases are depicted in Fig.\ref{fig:optimal_powersys_108} and Fig.\ref{fig:optimal_powersys_207}. Detailed values corresponding to these solutions can be referenced in Appendix~\ref{appendix:case_values}. For Case 1, attention matrices across three windows are presented in Fig.\ref{fig:node_attention_matrix_108}, while for Case 2, they are showcased in Fig.\ref{fig:node_attention_matrix_207}.

In Case 1, wind serves as the primary energy resource. Given the substantial capacity of wind generators, they are able to supply abundant energy. As a result, there's minimal reliance on conventional resources. Notably, most of the pivotal nodes are situated in Northern Europe. A few significant nodes were selected for analysis. Nodes 22 (Latvia) and 15 (Greece), highlighted in window 1, generate power predominantly from wind. Their strategic location, surrounded by energy-consuming nodes, ensures that the power they produce is directly consumed by adjacent nodes or those in close proximity. Other power-generating nodes include 9 (Denmark) and 27 (Poland) highlighted in window 2, and 25 (Netherlands) in window 3. Nodes 8 (Germany) and 14 (U.K.), highlighted in windows 2 and 3 respectively, predominantly consume power.

In Case 2, solar energy is the primary source. However, the available solar capacity is insufficient to meet the demand, leading to the activation of conventional generators. Here, significant nodes start emerging in Southern Europe. Nodes generating power via solar include 22 (Latvia), 33 (Slovak), 30 (Serbia), 19 (Italy), and 15 (Greece), as highlighted in windows 1 and 2. Nodes 26 (Norway) and 24 (North Macedonia), highlighted in window 3, are major power consumers.

\begin{figure*}[ht!]
    \centering
    \begin{subfigure}{.9\textwidth}
        \centering
        \includegraphics[width=\linewidth]{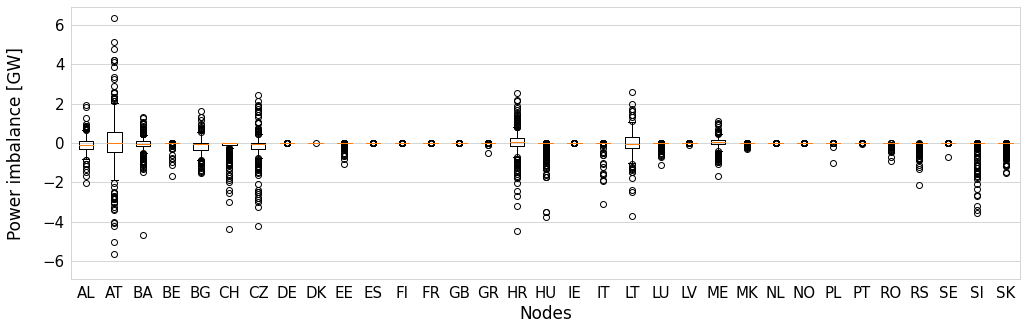}
        \caption{Before post-processing}
        \label{fig:postprocess_before}
    \end{subfigure}
    \par\medskip
    \begin{subfigure}{.9\textwidth}
        \centering
        \includegraphics[width=\linewidth]{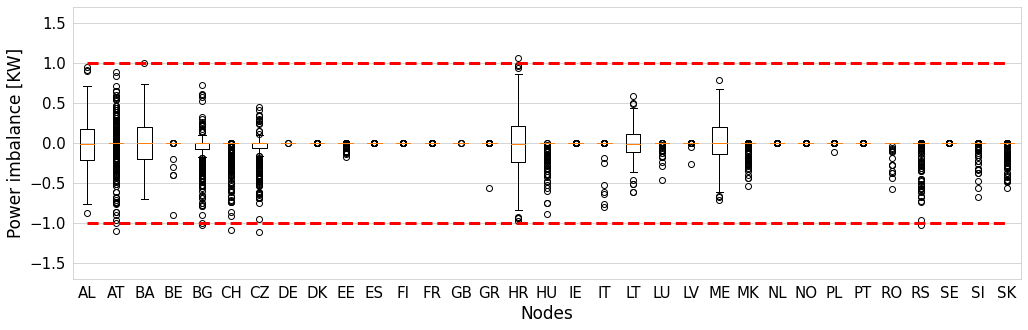}    
        \caption{After post-processing}
        \label{fig:postprocess_after}
    \end{subfigure} 
    \caption{Average power imbalance of predictions at each node.}
    \label{fig:postprocess}
\end{figure*}

\subsection{Accuracy of Prediction}
We begin by showcasing the efficacy of post-processing, which steers infeasible solutions into the feasible space. In Fig.\ref{fig:postprocess}, the average power imbalance for each node, both before and after post-processing, is computed over the test set. Nodes with positive imbalance values indicate a need for energy imports, while nodes with negative values suggest the contrary. As illustrated in Fig.\ref{fig:postprocess_before}, the imbalance magnitude prior to post-processing is in the GW range, with an average absolute imbalance across all nodes of 0.185 GW. However, post-processing rigorously directs infeasible solutions onto the boundary of the feasible space where no power imbalance is permitted. Consequently, as depicted in Fig.~\ref{fig:postprocess_after}, the imbalance magnitude after post-processing drops to the KW range, with an average absolute imbalance of just 0.454 KW. It's noteworthy that all experiments maintain a power precision of 1 KW, represented by the red dashed lines. After the post-processing projection, the majority of predictions fall within feasible bounds. Any slight deviations that arise are due to the projection's suboptimality.

\begin{figure*}[ht]
    \centering
    \begin{subfigure}{.9\textwidth}
        \centering
        \includegraphics[width=\linewidth]{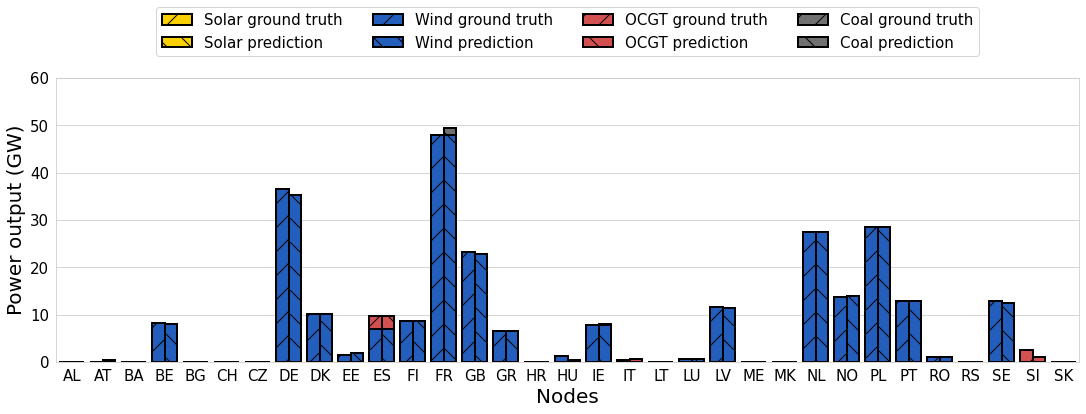}
        \caption{Case 1}
        \label{fig:optimal_powergen_108}
    \end{subfigure}
    \par\medskip
    \begin{subfigure}{.9\textwidth}
        \centering
        \includegraphics[width=\linewidth]{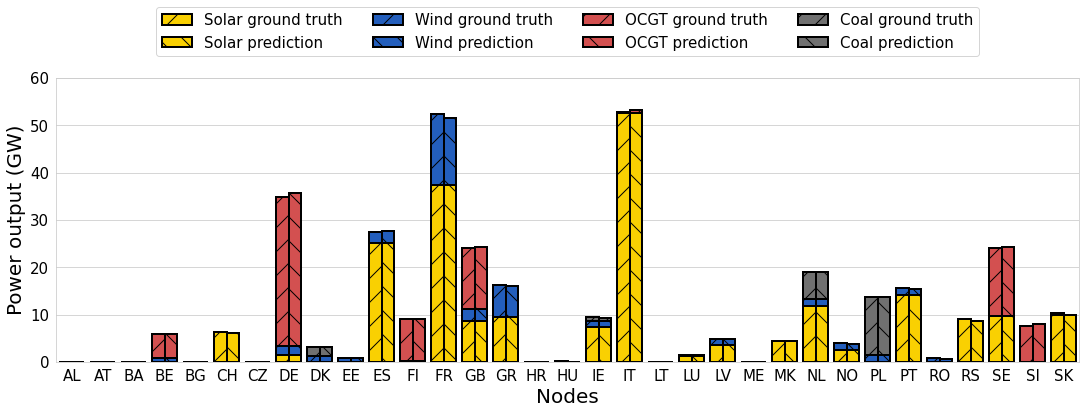}    
        \caption{Case 2}
        \label{fig:optimal_powergen_207}
    \end{subfigure} 
    \caption{Optimal power outputs of generators at each node.}
    \label{fig:optimal_power_output}
\end{figure*}

\begin{figure*}[ht!]
    \begin{center}
        \includegraphics[width=0.75\textwidth]{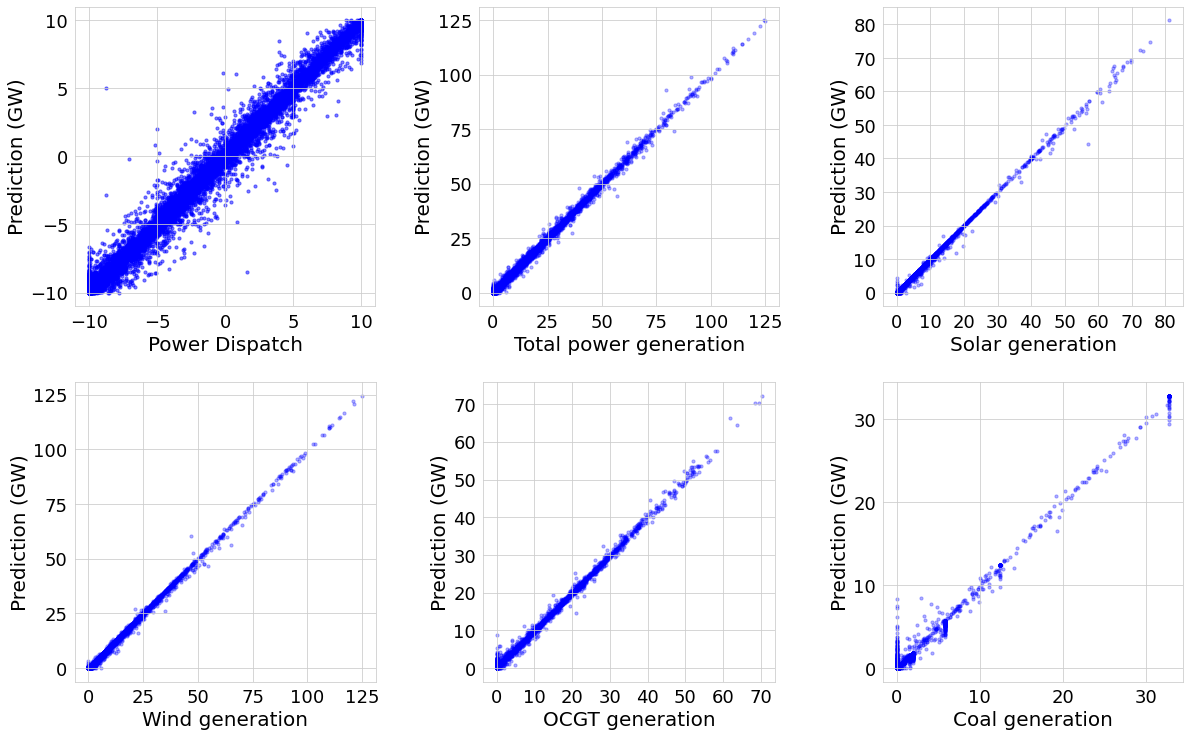}
    \end{center}
    \caption{Relationship between predictions and ground truth values for 33-nodes scenario.}
    \label{fig:statistical33}
\end{figure*}

\begin{figure*}[ht!]
    \begin{center}
        \includegraphics[width=0.75\textwidth]{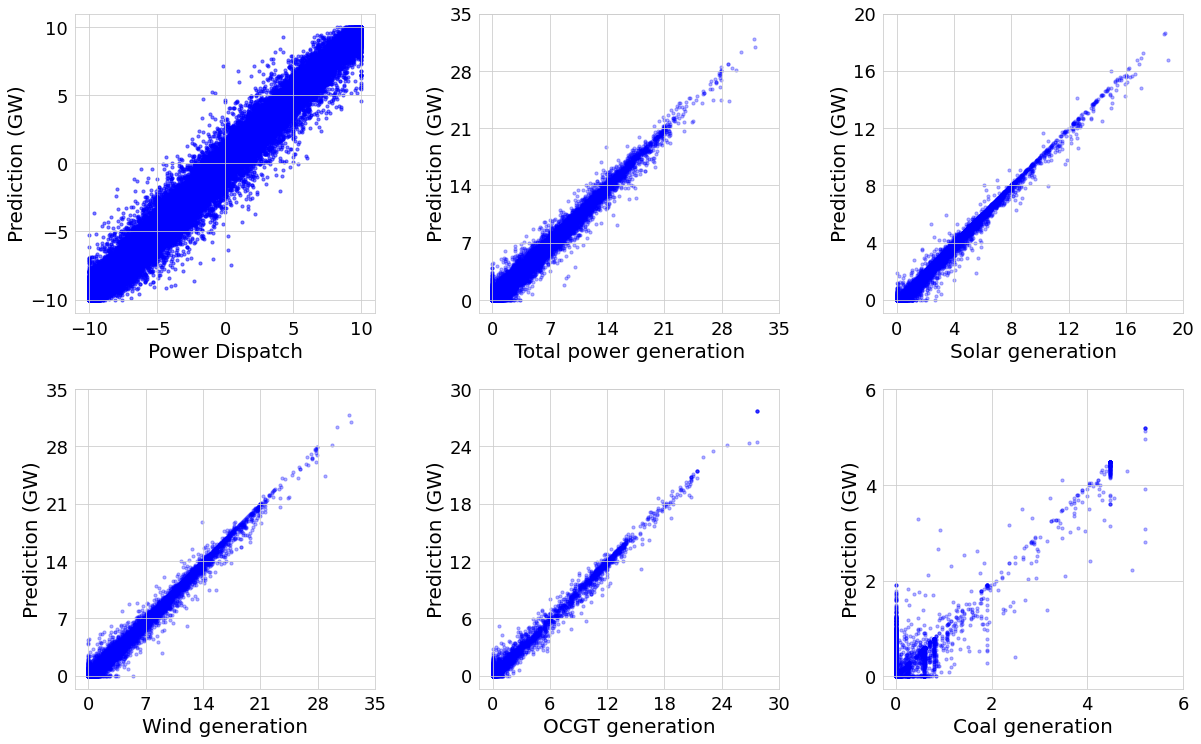}
    \end{center}
    \caption{Relationship between predictions and ground truth values for 300-nodes scenario.}
    \label{fig:statistical300}
\end{figure*}

We now present the optimal power outputs from each generator, broken down by energy source, for each node. Figures~\ref{fig:optimal_power_output} display the results for both Case 1 and Case 2. These predictions are juxtaposed with the ground truth values. On the horizontal axis, nodes are enumerated by their abbreviated codes, while the vertical axis quantifies the power outputs in GW. The power contributions from each generator at specific nodes are visualized as color-coded stacked bars. Notably, in the 33-node scenario, predictions closely mirror the ground truth. Given the visual complexity of representing absolute power outputs in the 300-node scenario, we chose to illustrate the correlation between predicted and actual values in Fig.\ref{fig:statistical33} for the 33-node scenario and Fig.\ref{fig:statistical300} for the 300-node scenario.

In Fig.~\ref{fig:statistical33}, the correlation between predicted and actual values for power dispatch across each link for the 33-node scenario, as well as the power output from different types of generators, is depicted using scatter plots. A robust proportional relationship is evident from these plots, which also provide insights into the output distribution of various energy sources. For instance, wind power dominates in terms of total energy capacity, with peak outputs reaching up to 125 GW. In contrast, OCGT and coal, serving as backup energy sources, possess relatively limited capacities.

In Fig.~\ref{fig:statistical300}, similar trends emerge for the 300-node scenario, although the variability in prediction errors widens due to the increased complexity of the scenario. The discrepancies are particularly pronounced for coal generators, attributed to their narrow output range and their designation as backup generators, which are typically activated as a last resort.

\subsection{Model Comparison}

\begin{figure*}[h]
    \centering
    \begin{subfigure}{.45\textwidth}
        \centering
        \includegraphics[width=\linewidth]{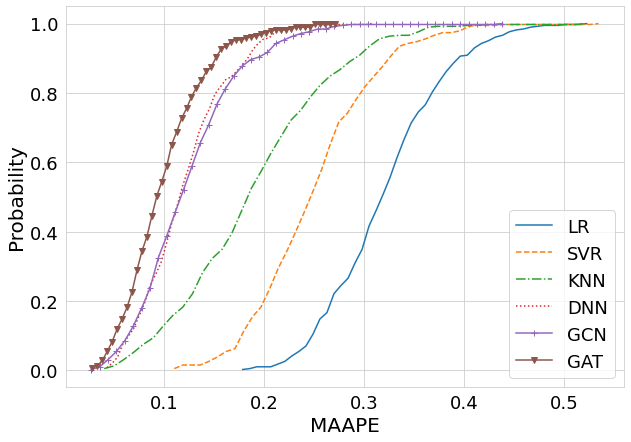}
        \caption{33-nodes scenario}
        \label{fig:accuracy_comparison_33}
    \end{subfigure}
    \hspace{-0.1cm}
    \begin{subfigure}{.45\textwidth}
        \centering
        \includegraphics[width=\linewidth]{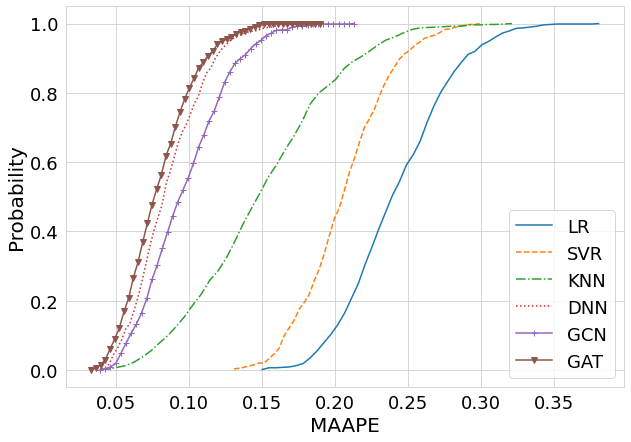}    
        \caption{300-nodes scenario}
        \label{fig:accuracy_comparison_300}
    \end{subfigure} 
    \caption{Accuracy comparison.}
    \label{fig:accuracy_comparison}
\end{figure*}

\begin{table*}[h]
    \centering
    \renewcommand\arraystretch{1.3}
    \caption{Runtime comparison (the runtime in a unit of sec./100 data points).}
    \begin{tabular}{@{}rrrrrrrr@{}}
        \toprule %[2pt]设置线宽
        \multirow{2}{*}{Scenarios} & \multicolumn{7}{c}{Runtime} \\
        \cline{2-8}
        \multirow{2}{*}{} & LR & SVR & KNN & DNN & GCN & GAT & IP \\
        \midrule %[2pt]  
        33 nodes & 0.01 & 2.00 & 1.19 & 1.71 & 1.29 & 2.40 & 2.60\\
        300 nodes & 0.12 & 149.26 & 50.00 & 4.06 & 5.39 & 4.90 & 22.22\\
        \bottomrule %[2pt]     
    \end{tabular}
    \label{tab:runtime_comparison}
\end{table*}

The proposed method excels in delivering not only interoperability but also highly accurate solutions. Fig.\ref{fig:accuracy_comparison} illustrates the cumulative distribution of Mean Arctangent Absolute Percentage Error (MAAPE) for both the 33-node scenario, as depicted in Fig.\ref{fig:accuracy_comparison_33}, and the 300-node scenario, as shown in Fig.\ref{fig:accuracy_comparison_300}. For comparison, we consider conventional methods—namely, Linear Regressor (LR), Support Vector Regressor (SVR), and k-Nearest Neighbors Regressor (KNN)—as well as neural network approaches, including Deep Neural Network (DNN) and Graph Convolutional Network (GCN). For the MAAPE metric, lower values signify better precision. Moreover, in terms of the cumulative distribution function, a steeper slope and closer proximity to the vertical axis signal superior accuracy. Note that the 300-node scenario utilized double the amount of training data compared to the 33-node scenario. This increase in training data is reflected in the enhanced accuracy observed across all data-driven methods in Fig.\ref{fig:accuracy_comparison_300}.

We provide a comparison of runtime performance in Table ~{\ref{tab:runtime_comparison}}. All data-driven models undergo initial training first and then are utilized to generate predictions for 100 data points. The performance of these models is compared with the Interior Point (IP) method, a conventional OPF solver, optimized by Gurobi using the same set of 100 data points, which is also listed in the table. As can be observed, the proposed Graph Attention Network (GAT) approach does not exhibit a significant runtime advantage in the context of small-scale power systems. However, its superiority in terms of runtime becomes distinctly evident in large-scale power systems—a trend that is similarly observed in the other neural network models.

\section{Conclusions} \label{Sec6}

In this study, we introduced a cutting-edge attention-based machine learning framework tailored to address the optimal power flow challenges inherent in predominantly renewable power systems. Our approach leverages the graph attention neural network, enabling the extraction of the attention matrix for each node within the power grid. Additionally, we employ an attention mechanism reminiscent of the transformer model to ascertain attention matrices for both nodes and connecting links. This graph attention based method adeptly discerns correlations, pinpointing pivotal nodes as influenced by diverse weather inputs. Furthermore, we crafted a unique machine learning strategy that seamlessly merges with projection post-processing, ensuring our algorithm consistently produces strictly feasible solutions.

In our comprehensive evaluations across two renewable power system scenarios, the efficacy of graph self-attention was ascertained through an in-depth exploration of the PCA analysis on the attention matrices. This not only elucidated the inner workings of neural networks but also laid the groundwork for more transparent AI interpretations. Our empirical analysis and assessment, encompassing two case studies, investigated the precision of predictions and their alignment with ground truths. The dispersion of average power imbalances attested to the efficiency of our post-processing and underscored our method's aptitude in delivering feasible solutions. Notably, besides the interpretability, our approach demonstrated superior performance surpassing existing data-driven methodologies in the field.

The integration of machine learning (ML)-based optimal power dispatch solutions in real-time heralds a transformative era for modern power systems, offering a trifecta of enhanced efficiency, reliability, and sustainability. These advanced solutions are poised to deliver real-time decision support and catalyze the seamless integration of distributed energy resources. Beyond these capabilities, they are strategically designed to enable intelligent microgrid management, ensuring stringent frequency regulation, and providing a robust framework for economic optimization. This is achieved by taking into account a comprehensive range of physical factors in conjunction with the prevailing dynamics of the energy market. Parallelly, the graph attention mechanism, a cornerstone of this approach, unveils a capacity for neural network interpretation, thereby paving the way for increasingly precise solutions. This heightened precision is particularly invaluable when confronting intricate challenges, such as the orchestration of virtual power plants, further illustrating the broad potential of this innovative framework.

Future work may look into a deeper exploration of the mechanisms underlying attention matrices, aiming to further enhance the interpretability of neural networks. A promising direction involves the incorporation of storage units, given their pivotal role in the context of future energy conversion systems. In addition, the examination of more sophisticated machine learning algorithms is warranted, particularly for addressing the complex OPF problem where variables exhibit both spatial and temporal correlations. As the integration of renewable energy sources continues to escalate, it is essential to assimilate the uncertainties inherent in their outputs within a stochastic OPF framework. This would facilitate more effective management of variable renewable resources. Moreover, there is a promising opportunity to extend the application of ML-based OPF to distributed power systems. Such applications would necessitate coordination with neighboring subsystems and should ideally encompass considerations from both the energy market and demand-side management, thereby offering a more holistic and responsive solution.

\section*{Acknowledgement}
This work was funded by the Xidian-FIAS International Joint Research Center (XF-IJRC). The responsibility for the contents lies solely with the authors.

\begin{appendices}
\section{Power system details}\label{appendix:para}
\begin{table*}[h!]
    \centering
    \renewcommand\arraystretch{1.3}
    \caption{Settings for different type of generators.}
    \begin{tabular}{@{}rrrrrrr@{}}
        \toprule %[2pt]设置线宽     
        Generators & \makecell[c]{Capacity\\(initial)} & \makecell[c]{Capital cost\\(currency/MW)} & \makecell[c]{Marginal cost\\(currency/MWh)} & Efficiency & \makecell[c]{$\textrm{CO}_{2}$ Emission\\(ton/MWh)} & \makecell[c]{Actual Marginal cost\\(currency/MWh)} \\
        \midrule %[2pt]  
        Coal & $>$0 & 145,000 & 25.000 & 0.33 & 1.0 & 125.00\\
        OCGT & $>$0 & 49,400 & 58.385 & 0.41 & 0.635 & 121.89\\
        Wind & 0 & 127,450 & 0.015 & 1.0 & 0.0 & 0.015\\
        Solar & 0 & 61,550 & 0.010 & 1.0 & 0.0 & 0.010\\
        \bottomrule %[2pt]     
    \end{tabular}
    \label{tab:gen_set}
\end{table*}

\begin{figure*}[ht!]
    \begin{center}
        \begin{subfigure}{0.45\textwidth}
        \includegraphics[width=\linewidth]{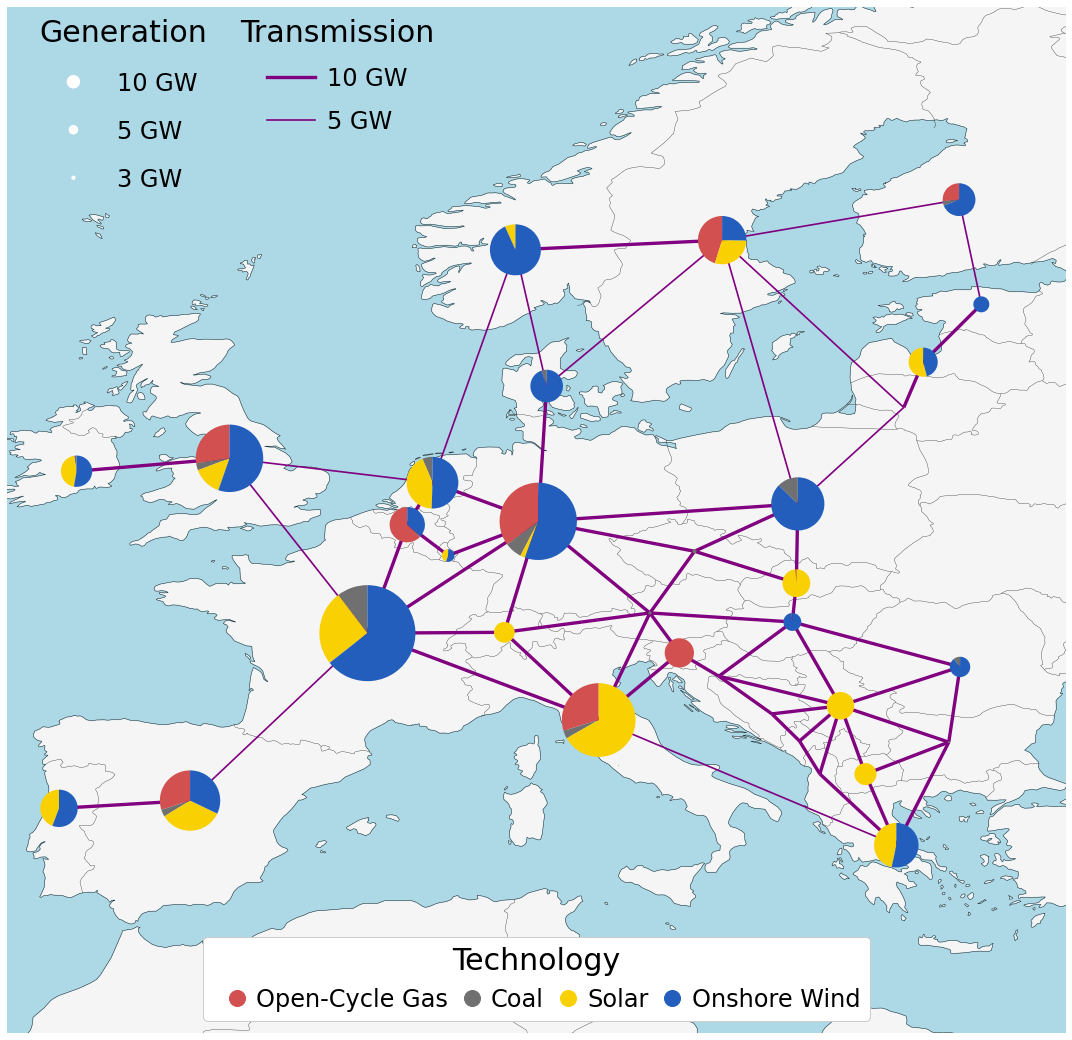}
        \caption{} 
        \label{fig:33_sys_cap_a}
        \end{subfigure}%
        \hspace*{0cm}   
        \begin{subfigure}{0.45\textwidth}
        \includegraphics[width=\linewidth]{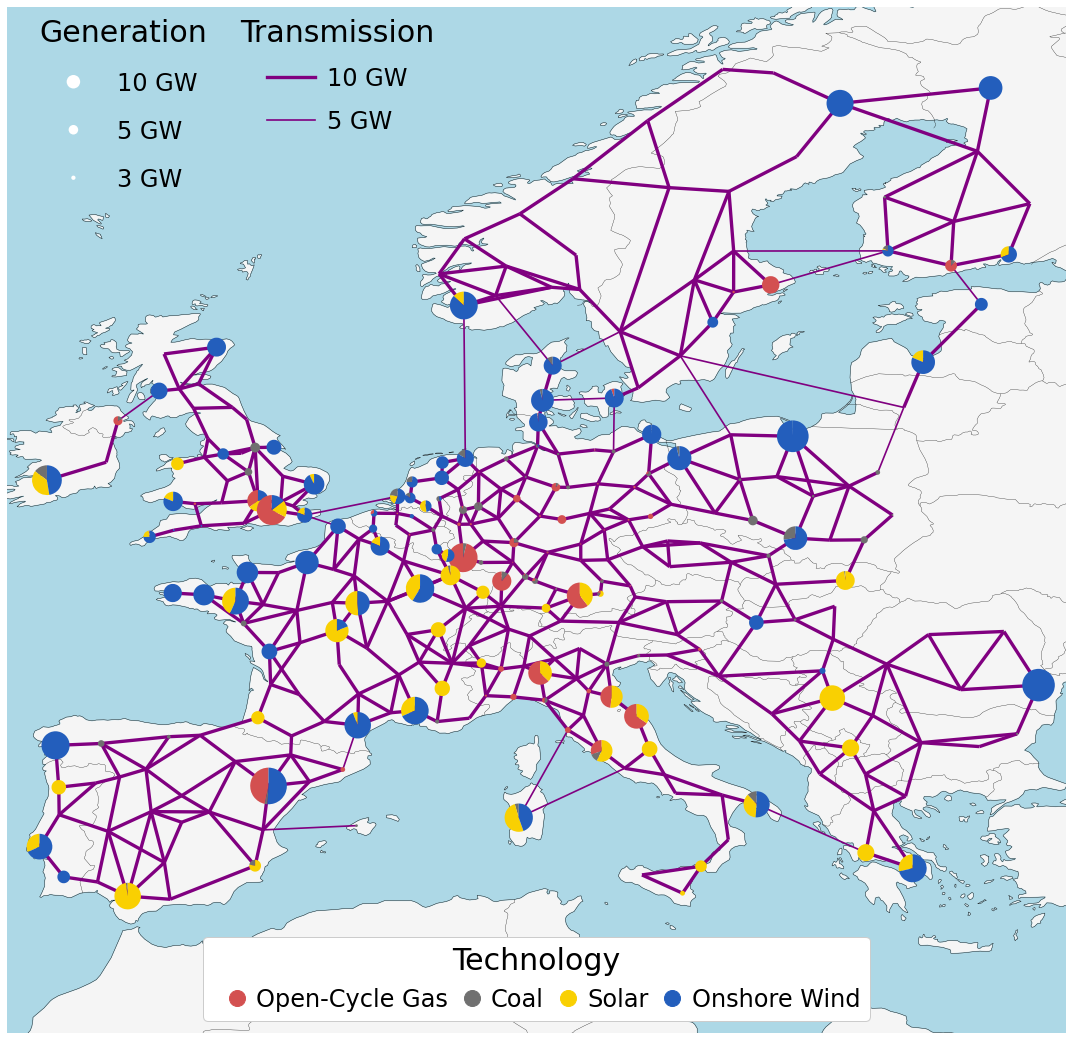}
        \caption{} 
        \label{fig:300_sys_cap_b}
        \end{subfigure}
    \end{center}
    \caption{Energy capacity of clustered power system.}
    \label{fig:two_sys_wcap}
\end{figure*}

\begin{figure*}[h!]
    \begin{center}
        \begin{subfigure}{6cm}
        \includegraphics[width=\linewidth]{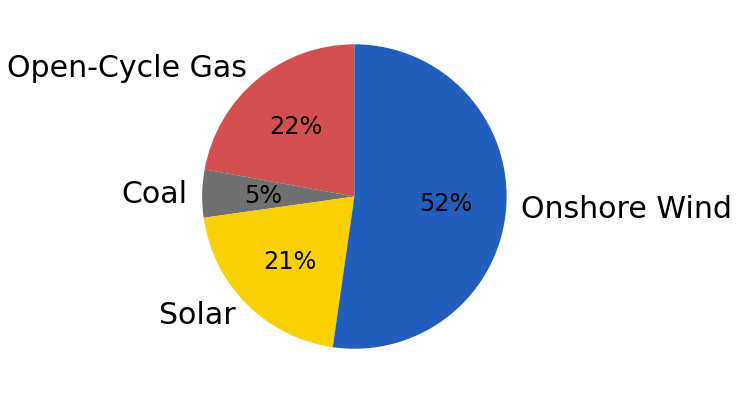}
        \caption{} 
        \label{fig:cap_pie_a}
        \end{subfigure}%
        \hspace*{-0.4cm}   
        \begin{subfigure}{5.5cm}
        \includegraphics[width=\linewidth]
        {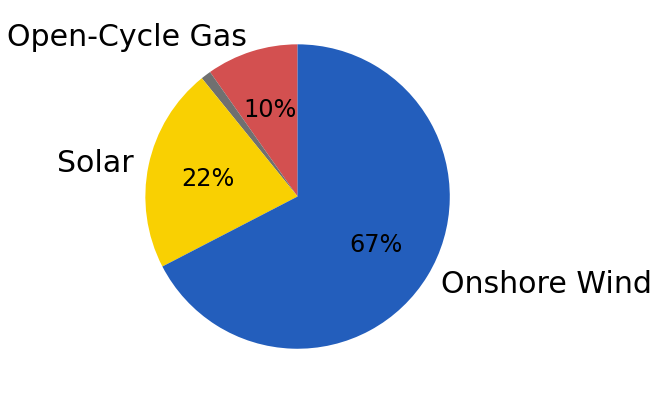}
        \caption{} 
        \label{fig:cap_pie_b}
        \end{subfigure}
    \end{center}
    \caption{Proportion for each energy type.}
    \label{fig:capacity_pie}
\end{figure*}

Table \ref{tab:gen_set} shows the parameters for generators with different carriers. Before optimization, the initial capacity of renewable generator is set to be 0, however the initial capacity of conventional generator is set to be positive value. Since the capacity of each generator can only be increased, this ensures a minimum capacity of conventional generators. If the carbon price is considered, e.g., 100 (currency/ton), the actual marginal cost of generator with different energy carrier is the sum of it's marginal cost and $\textrm{CO}_{2}$ emission cost which is the product of carbon price and $\textrm{CO}_{2}$ emission amount. Other parameters are collected from \cite{schlott2021carbon}.

For links between each pair of nodes, they came from two kinds of component, we set both of them as controllable directed power links. One kind came from the passively determined transmission lines yielding after clustering, with the nominal active power of 10000 MW. The other kind came from controllable directed power links, with the nominal active power of 5000 MW. The nominal active power is the limit of active power that can pass through the link, which is set arbitrary and identical for each kind, just for simplification. The final controllable links have the efficiency of 0.9 and marginal cost of 3.642 (currency/MWh) \cite{marginalcost2004}.

For input data, the electricity consumption data we used were originated from the open source website, European Network of Transmission System Operators for Electricity (ENTSO-E) \cite{entsoe}. And the weather data were originated from the public dataset ERA5 by the European Centre for Medium-Range Weather Forecasts (ECMWF) \cite{era5}, then the renewable generation capacity index time series for wind and solar are derived by PyPSA-EUR. For both electricity consumption and weather data, we took the year 2013 and 2014 arbitrarily, since different year's data will not influence the trend and analysis of results. There are 8760 data points in hourly resolution each year, and totally 17520 data points for two years.

\begin{table*}[ht!]
    \centering
    \renewcommand\arraystretch{1.3}
    \caption{Node number and corresponding country name in 33 nodes scenario.}
    \begin{tabular}{@{}lll|lll|lll@{}}
        \toprule %[2pt]设置线宽     
        No. & Country & Short & No. & Country & Short & No. & County & Short \\ 
        \midrule %[2pt] 
        1 & Albania & AL & 12 & Finland & FI & 23 & Montenegro & ME \\
        2 & Austria & AT & 13 & France & FR & 24 & North Macedonia & MK \\
        3 & Bosnia and Herzegovina & BA & 14 & United Kingdom & GB & 25 & Netherlands & NL \\
        4 & Belgium & BE & 15 & Greece & GR & 26 & Norway & NO \\
        5 & Bulgaria & BG & 16 & Croatia & HR & 27 & Poland & PL \\
        6 & Switzerland & CH & 17 & Hungary & HU & 28 & Portugal & PT \\
        7 & Czech Republic & CZ & 18 & Ireland & IE & 29 & Romania & RO \\
        8 & Germany & DE & 19 & Italy & IT & 30 & Serbia & RS \\
        9 & Denmark & DK & 20 & Lithuania & LT & 31 & Sweden & SE \\
        10 & Estonia & EE & 21 & Luxembourg & LU & 32 & Slovenia & SI \\
        11 & Spain & ES & 22 & Latvia & LV & 33 & Slovak Republic & SK \\
        \bottomrule %[2pt]     
    \end{tabular}
    \label{tab:no_country}
\end{table*}

\begin{table*}[h!]
    \centering
    \renewcommand\arraystretch{1.3}
    \caption{Link number and nodes that connected in 33 nodes scenario.}
    \begin{tabular}{@{}lrr|lrr|lrr|lrr@{}}
        \toprule
        Link & $\textrm{Node}_0$ & $\textrm{Node}_1$ & Link & $\textrm{Node}_0$ & $\textrm{Node}_1$ & Link & $\textrm{Node}_0$ & $\textrm{Node}_1$ & Link & $\textrm{Node}_0$ & $\textrm{Node}_1$ \\ \midrule
        1 & 19 & 15 & 16 & 26 & 31 & 31 & 8 & 21 & 46 & 11 & 28 \\
        2 & 12 & 10 & 17 & 3 & 30 & 32 & 1 & 30 & 47 & 3 & 16 \\
        3 & 13 & 11 & 18 & 1 & 23 & 33 & 4 & 13 & 48 & 4 & 25 \\
        4 & 9 & 26 & 19 & 8 & 9 & 34 & 20 & 22 & 49 & 2 & 8 \\
        5 & 27 & 20 & 20 & 17 & 30 & 35 & 17 & 29 & 50 & 5 & 15 \\
        6 & 31 & 12 & 21 & 4 & 21 & 36 & 27 & 33 & 51 & 17 & 33 \\
        7 & 27 & 31 & 22 & 5 & 30 & 37 & 16 & 32 & 52 & 23 & 30 \\
        8 & 25 & 26 & 23 & 13 & 19 & 38 & 16 & 17 & 53 & 8 & 25 \\
        9 & 14 & 25 & 24 & 6 & 13 & 39 & 5 & 29 & 54 & 3 & 23 \\
        10 & 20 & 31 & 25 & 6 & 8 & 40 & 2 & 19 & 55 & 16 & 30 \\
        11 & 14 & 13 & 26 & 2 & 6 & 41 & 10 & 22 & 56 & 7 & 27 \\
        12 & 31 & 9 & 27 & 7 & 33 & 42 & 5 & 24 & 57 & 2 & 7 \\
        13 & 6 & 19 & 28 & 8 & 27 & 43 & 14 & 18 & 58 & 1 & 15 \\
        14 & 7 & 8 & 29 & 29 & 30 & 44 & 15 & 24 & 59 & 2 & 32 \\
        15 & 19 & 32 & 30 & 24 & 30 & 45 & 8 & 13 & 60 & 2 & 17 \\ \bottomrule
\end{tabular}
    \label{tab:link_nodes}
\end{table*}

\section{Structure of clustered power grids}
\label{appendix:cluster}
After clustering, we got two power system models with 33 nodes and 300 nodes respectively. Fig.~\ref{fig:two_sys_wcap} shows the energy capacity at each nodes, there are 4 kinds of technologies in total. Both in Fig.~\ref{fig:33_sys_cap_a} the 33 nodes network and Fig.~\ref{fig:300_sys_cap_b} the 300 nodes network, we can see that wind generators are more deployed in northern Europe than in southern part due to the climate, and also are located more commonly along the coastline. However solar generators are more deployed in southern Europe than northern part due to the sufficient sunshine. Conventional generators are more deployed in the central Europe, since nodes in central Europe don't have neither much wind nor solar resources, even some of the inland nodes can acquire energy just by importing as long as there are enough transmission capacities. Fig.~\ref{fig:capacity_pie} shows the proportion of different energy resources in the whole system, wind energy shares the most in the both two models. Since we analyzed 33-nodes power grid in the most cases, and in that case each country is denoted by a node, the comparison of node numbers and countries is provided in Table~\ref{tab:no_country}. Then links in the power grid are shown in Fig.~\ref{tab:link_nodes}, as well as nodes that are connected by the links.

\begin{table*}[h]
    \centering
    \renewcommand\arraystretch{1.3}
    \caption{Power generation, demand and export situation for Case 1.}
    \begin{tabular}{@{}rrrr|rrrr|rrrr@{}}
        \toprule
        Node & Generation & Demand & Export & Node & Generation & Demand & Export & Node & Generation & Demand & Export \\ \midrule
        1 & 0.0 & 449.9 & -449.9 & 12 & 8693.4 & 9193.4 & -500.0 & 23 & 0.0 & 297.0 & -297.0 \\
        2 & 0.0 & 5510.5 & -5510.5 & 13 & 48011.0 & 44825.2 & 3185.9 & 24 & 0.0 & 812.0 & -812.0 \\
        3 & 0.0 & 1001.0 & -1001.0 & 14 & 23142.5 & 28957.4 & -5814.9 & 25 & 27474.0 & 9974.0 & 17500.0 \\
        4 & 8280.5 & 8446.0 & -165.5 & 15 & 6585.0 & 4031.0 & 2554.0 & 26 & 13618.5 & 12812.6 & 805.9 \\
        5 & 0.0 & 3458.6 & -3458.6 & 16 & 0.0 & 1383.0 & -1383.0 & 27 & 28577.1 & 13521.0 & 15056.1 \\
        6 & 0.0 & 3713.2 & -3713.2 & 17 & 1159.6 & 3726.0 & -2566.4 & 28 & 12891.6 & 4753.0 & 8138.6 \\
        7 & 0.0 & 5939.0 & -5939.0 & 18 & 7885.6 & 2652.1 & 5233.4 & 29 & 931.8 & 4984.0 & -4052.2 \\
        8 & 36440.8 & 47716.1 & -11275.4 & 19 & 445.7 & 26965.2 & -26519.5 & 30 & 0.0 & 3465.0 & -3465.0 \\
        9 & 10084.7 & 1584.7 & 8500.0 & 20 & 0.0 & 901.0 & -901.0 & 31 & 12797.3 & 13724.0 & -926.7 \\
        10 & 1479.5 & 744.0 & 735.5 & 21 & 616.7 & 703.0 & -86.3 & 32 & 2492.2 & 1692.5 & 799.7 \\
        11 & 9625.4 & 22634.8 & -13009.4 & 22 & 11647.3 & 597.0 & 11050.3 & 33 & 0.0 & 2671.0 & -2671.0 \\ \bottomrule
    \end{tabular}
    \label{tab:GDE_for_nodes_case1}
\end{table*}

\begin{table*}[h]
    \centering
    \renewcommand\arraystretch{1.3}
    \caption{Power generation, demand and export situation for Case 2.}
    \begin{tabular}{@{}rrrr|rrrr|rrrr@{}}
        \toprule
        Node & Generation & Demand & Export & Node & Generation & Demand & Export & Node & Generation & Demand & Export \\ \midrule
        1 & 0.0 & 495.3 & -495.3 & 12 & 9052.5 & 8674.9 & 377.5 & 23 & 0.0 & 465.0 & -465.0 \\
        2 & 0.0 & 7859.6 & -7859.6 & 13 & 52420.8 & 50611.1 & 1809.8 & 24 & 4436.0 & 894.0 & 3542.0 \\
        3 & 0.0 & 1523.0 & -1523.0 & 14 & 24078.8 & 37951.4 & -13872.6 & 25 & 19066.3 & 14533.0 & 4533.3 \\
        4 & 5868.1 & 9716.0 & -3847.9 & 15 & 16315.0 & 7196.0 & 9119.0 & 26 & 3888.0 & 10604.1 & -6716.1 \\
        5 & 0.0 & 4280.8 & -4280.8 & 16 & 0.0 & 2355.0 & -2355.0 & 27 & 13761.6 & 17801.0 & -4039.4 \\
        6 & 6252.6 & 4428.2 & 1824.4 & 17 & 107.1 & 5048.0 & -4940.9 & 28 & 15517.0 & 6517.0 & 9000.0 \\
        7 & 0.0 & 7044.0 & -7044.0 & 18 & 9382.3 & 3226.4 & 6155.9 & 29 & 785.5 & 5909.0 & -5123.5 \\
        8 & 34770.4 & 64666.5 & -29896.1 & 19 & 52927.4 & 49650.5 & 3276.9 & 30 & 9002.3 & 4283.0 & 4719.3 \\
        9 & 3174.0 & 2085.4 & 1088.7 & 20 & 0.0 & 1208.0 & -1208.0 & 31 & 24113.2 & 12448.0 & 11665.2 \\
        10 & 798.2 & 813.0 & -14.8 & 21 & 1445.4 & 898.8 & 546.6 & 32 & 7641.3 & 2089.4 & 5551.9 \\
        11 & 27509.5 & 33009.5 & -5500.0 & 22 & 4771.7 & 782.0 & 3989.7 & 33 & 10219.1 & 3129.0 & 7090.1 \\ \bottomrule
    \end{tabular}
    \label{tab:GDE_for_nodes_case2}
\end{table*}

\section{Hyperparameters}\label{appendix:hyper_para}
For all the power values, we set the precision to be $\delta=1$ KW, e.g. we ignored the generator whose nominal power is less than $\delta$. The length $m$ of positional encoding for nodes is 8 for 33-nodes case and 16 for 300-nodes case. 

Inside the SMW-GSAT layer, the dimension of latent node state $F'=64$, and we used 3 windows in total, for the 33-nodes case, each mask in the 3 windows focuses on 1-hop, 3-hop and 5-hop neighborhoods respectively, for the 300-nodes case, each mask focuses on 4-hop, 12-hop and 20-hop neighborhoods respectively. Inside NLAT layer, the dimension of intermediate value for quires and keys $V=64$, the number of latent features for each link $U=128$. The hidden units in MLP layer is $d_{k}=32,\; k=1,2,...,K$, number of hidden layers $K=2$. Hyperparameters $U$ and $K$ are set to consider improving the nonlinearity of the neural network architecture, but not introducing too much complexity. The training set contains 95\% of all the data points, and test set contains the rest 5\% data. The training was up to 1000 epochs with batch size 32. Early stopping strategy was applied with the tolerance of 100. The model was trained with stochastic gradient descent via Adam optimizer, the learning rate refers to the polynomial decay schedule from maximum $1e^{-3}$ to minimum $1e^{-4}$ with the decay step 100 and polynomial power 1.5. The scaling factor is $\alpha=1e^{-7}$.

For those methods used for comparison, we used default parameters provided by scikit-learn package to train LR, SVR, KNN and GPR model. The input for these methods is just the reshaped feature matrix, which mean there is no positional encoding. For DNN model, there were 4 hidden layers with 1024 hidden units each, dropout strategy was applied after each hidden layer with the dropout rate 0.2. For GCN+MLP model, the length of latent state of nodes was 64 and the size of layers in GCN was 4, MLP after GCN had 2 hidden layers with 1024 hidden units each, dropout strategy was also applied after each hidden layer with the dropout rate 0.2. The scaling factor used while training DNN and GCN+MLP is $\alpha=5e^{-5}$, the learning rate was same to what used while training the proposed network.

\section{Variable values for test cases}
\label{appendix:case_values}
Table~\ref{tab:GDE_for_nodes_case1} and Table~\ref{tab:GDE_for_nodes_case2} present the values of power generation, power consumption and exported power at each node for Case 1 and Case 2 respectively. Negative value for export indicates power consumed as that node is greater than power generated, and it is actually importing power from other nodes. Positive value indicates exporting.

\end{appendices}

\bibliographystyle{model1-num-names}
% \bibliographystyle{cas-model2-names}

\bibliography{cas-refs}

\end{document}